\documentclass{egpubl}
\usepackage{algpseudocode}
\usepackage{algorithm}
\usepackage{setspace}
\usepackage{multirow}
\usepackage{multicol}
\usepackage{amsmath}
\usepackage{booktabs}
\algnewcommand{\LeftComment}[1]{\Statex \(\triangleright\) #1}
 \electronicVersion 

\ifpdf \usepackage[pdftex]{graphicx}
\else \usepackage[dvips]{graphicx} \fi

\PrintedOrElectronic

\usepackage{t1enc,dfadobe}

\usepackage{egweblnk}
\usepackage{cite}

\title {Point cloud segmentation using hierarchical tree for architectural models}

\author{O.Hassaan, A.Shamail, Z.Butt, M.Taj \\
  Department of Computer Science, Syed Babar Ali School of Science and Engineering \\ Lahore University of Management Sciences, Pakistan\\
  omair.hassaan@lums.edu.pk, abeera.shamail@gmail.com, m.zainbutt@hotmail.com, murtaza.taj@lums.edu.pk}

\begin{document}

 \teaser{
	\begin{center}\leavevmode \centerline{
		\begin{tabular}{@{}c@{}c@{}c@{}c@{}}   
				\includegraphics[width=.53\columnwidth]{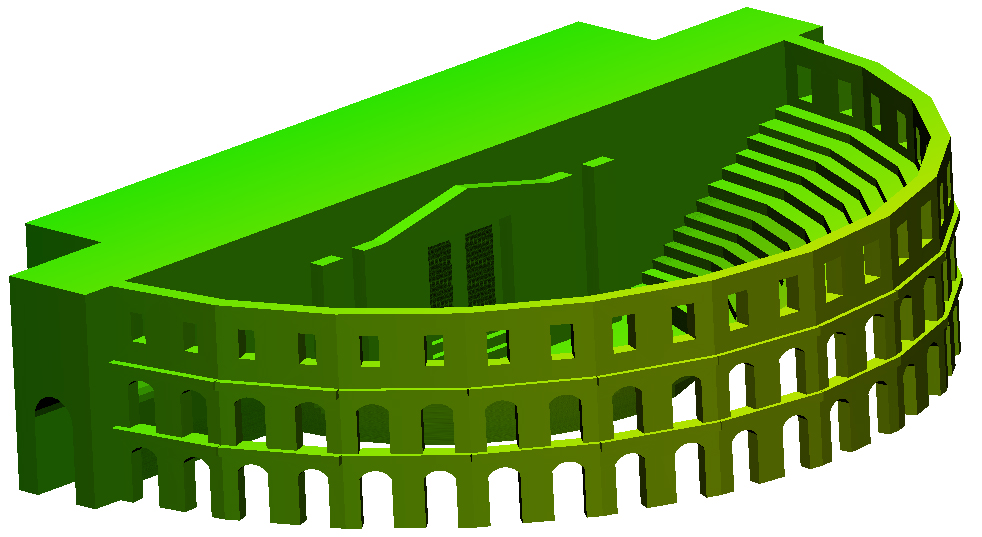} &	
        		\includegraphics[width=.53\columnwidth]{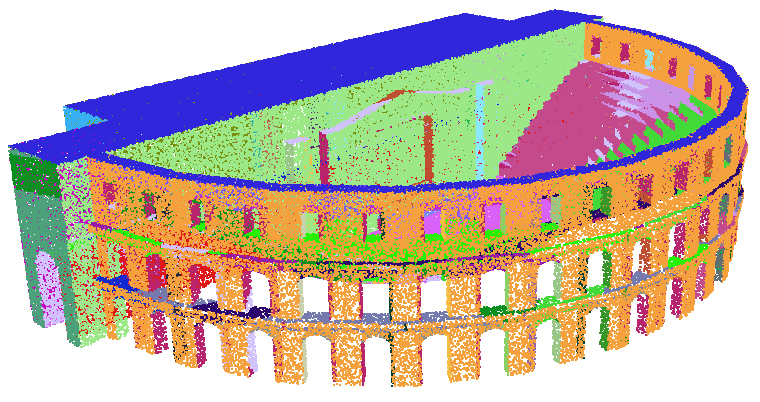} &
				\includegraphics[width=.5\columnwidth]{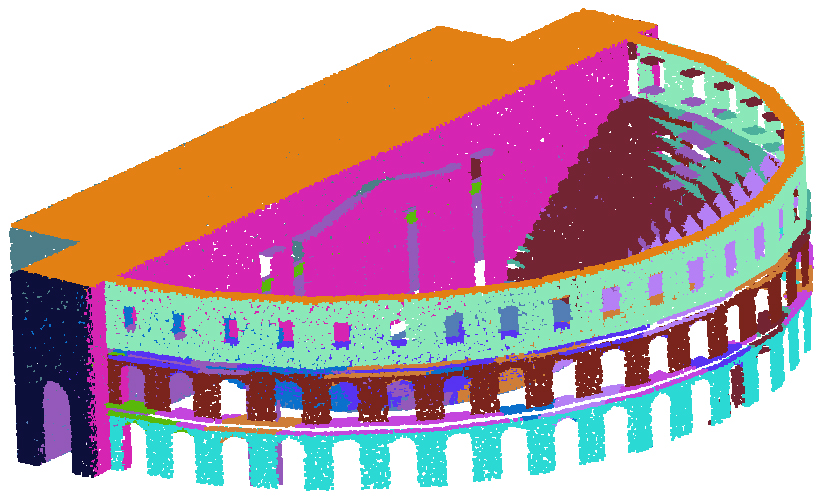} &
				\includegraphics[width=.5\columnwidth]{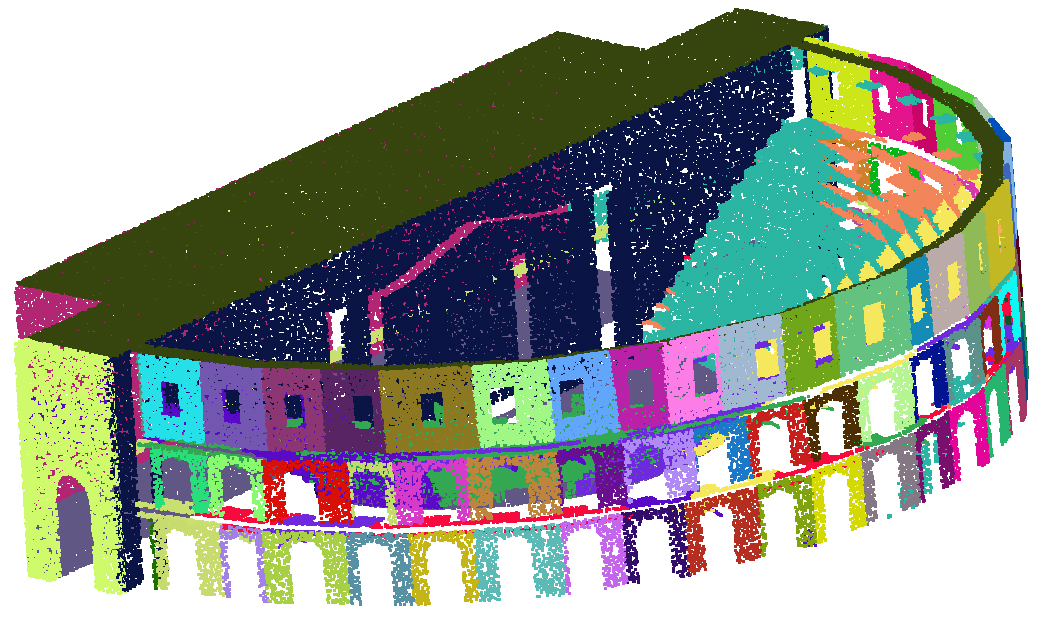}\\
 				(a) & (b) & (c) & (d) \\
		\end{tabular}
	} \caption{\label{fig:teaser} Example showing automatically generated detailed segmentation of raw point cloud. (a) Point cloud (b) Coarse segments using RANSAC. (c) DBScan Clustering. (d) Ours hierarchical segmentation.}
    \end{center}  
 \label{fig:teaserimg}
 }

\maketitle

\begin{abstract}
Recent developments in the 3D scanning technologies have made the generation of highly accurate 3D point clouds relatively easy but the segmentation of these point clouds remains a challenging area. A number of techniques have set precedent of either planar or primitive based segmentation in literature. In this work, we present a novel and an effective primitive based point cloud segmentation algorithm. The primary focus, i.e. the main technical contribution of our method is a hierarchical tree which iteratively divides the point cloud into segments. This tree uses an exclusive energy function and a 3D convolutional neural network, HollowNets to classify the segments. We test the efficacy of our proposed approach using both real and synthetic data obtaining an accuracy greater than 90\% for domes and minarets.  


\end{abstract}

\section{Introduction}
The recent advancements in 3D scanning technologies have paved way for the generation of highly accurate point cloud data. The increasing use of terrestrial and aerial laser scanning along with photogrammetry has enabled mapping of large-scale outdoor scenes, such as an entire housing block~\cite{Lin2013ToG} or a coarse model of a city~\cite{LafargeIJCV2012}. This new trend has paved way for emerging applications such as drone-based delivery of goods and self-driving cars that require detailed 3D understanding of large-scale outdoor scenarios. 3D scene understanding requires segmentation of point cloud into meaningful subsegments, however, segmentation of 3D point cloud into semantically consistent segments (see Fig~\ref{fig:realresults}) is still a challenging problem.

One of the seminal work in large-scale 3D point cloud manipulation is by Armeni~\cite{armeniCVPR16}. Their algorithm segments a scene using strong geometric priors for space estimation and then performs spatial parsing to segment known structures. They implement segmentation on architecturally similar buildings having significant number of known features. However, their work cannot be generalized to scenes with irregular objects and is confined to indoor scene segmentation only. 

Object recognition is often used in both 2D and 3D data to support the segmentation process. Martinovic~\cite{MartinovicCVPR2015} proposes one such approach that combines image based classification with 3D object classification for point cloud segmentation. Such approaches are not applicable to point cloud data in general as a dense set of 2D photographs is not always available as in the case of laser scanning. Similarly, hierarchical semantic segmentation ~\cite{Dohan:2015:LHS} is based on learning a Merge Classifier that predicts whether a combination of segments belongs to the same object instance or not. This is a bottom-up approach which suffers from combinatorial complexity.

\begin{figure*}[t]
	\begin{center}\leavevmode \centerline{
		\begin{tabular}{@{}c@{}}    
				\includegraphics[width=.5\textwidth]{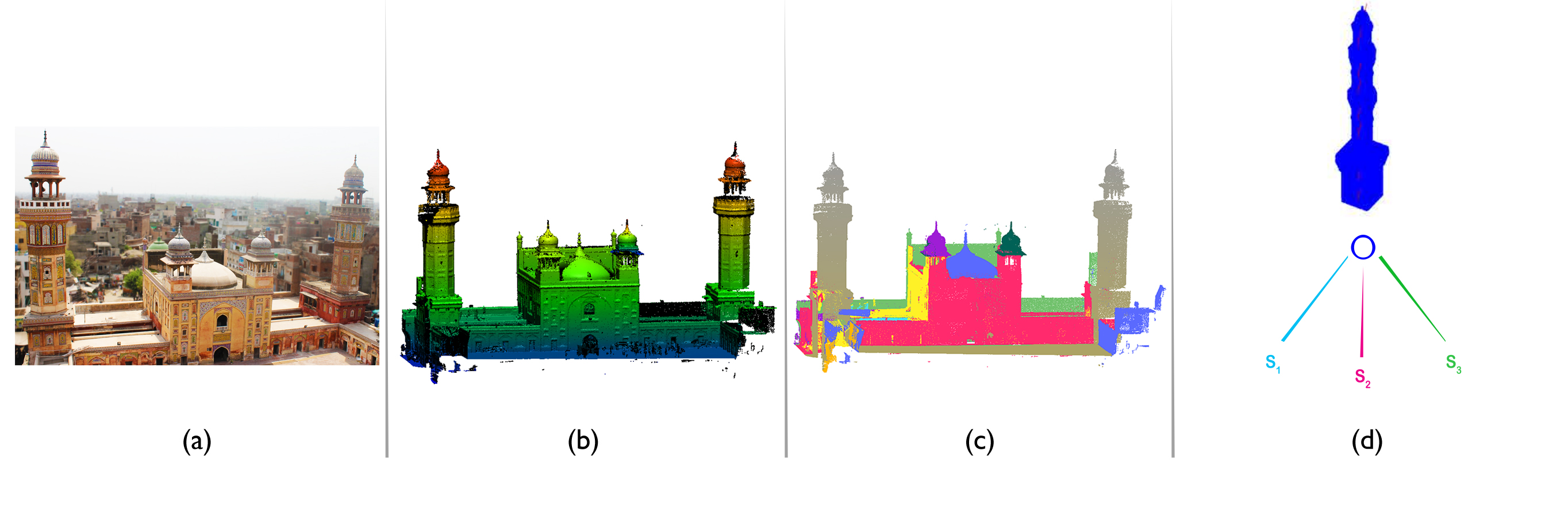}
        		\includegraphics[width=.5\textwidth]{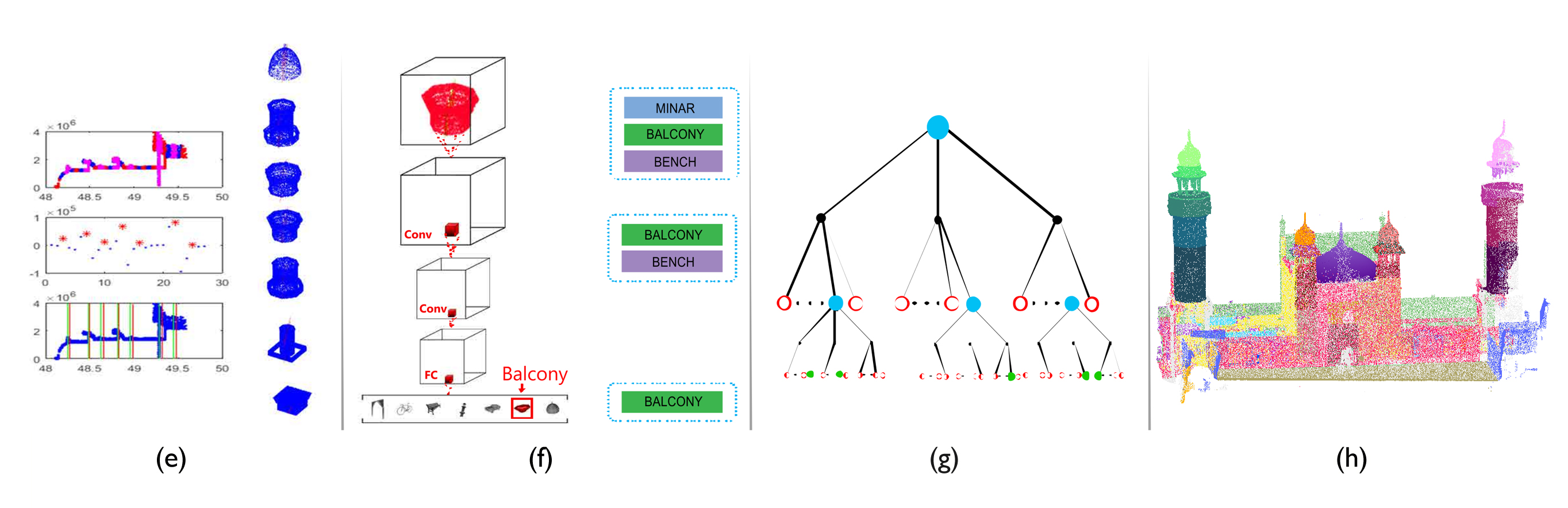} \\									                
		\end{tabular}
	} \caption{\label{fig:flow}
	Flow diagram of the proposed approach. (a) Input point cloud. (b) Coarse segments obtained using RANSAC~\cite{schnabel-2007-efficient} and spatial clustering. (c) Projection sequence. (d) Peak finding on profile curve of minar showing $7$ peaks as asterix and vertical line. (e) Classification of one of the segments. (f) Illustration of tree being updated as more segments being added. (g) Obtained segments.}
    \end{center}  
\end{figure*}

Lin et al.~\cite{Lin2013ToG} uses LiDAR data on low-rise houses using planar primitives, patches and symmetric blocks to segment a point cloud. This approach is confined to symmetric houses and planar surfaces and therefore, can not be generalized. Similarly, RAPter (RAPter)~\cite{MonszpartEtAl:RAPter:2015} exploits regular arrangement of planes to obtain multiple planar segments from a point cloud. 

Bajwa et al.~\cite{BajwaCGI2016} proposes an interactive coarse-to-fine segmentation approach using three fundamental Manhattan World constraints. They performe 1D and 2D projection of point cloud along the orthogonal planes and obtaine segments using peak finding on these projections. In their work, the projection applied on each point cloud segment is selected manually. Ours is a fully automatic solution based on hierarchal tree segmentation that automatically uses a combination of these projections and perform automatic segmentation.

Various techniques exist for the segmentation and classification of point cloud which use low level as well as high level constraints. ~\cite{grilli2017review} et. al. reviewed such techniques e.g edge based, region growing, segmentation by model fitting and machine learning based segmentation. Edge based and region growing algorithms are sensitive to noise and uneven density of point clouds. The most widely used algorithm employed in model fitting based category is Random Sample Consensus (RANSAC) ~\cite{fischler1981random}. RANSAC randomly draws minimal data points to construct shape primitives and determines the best fit. Although RANSAC based algorithms generally perform well but fall short on complex structures e.g in architectural fields details can not be modeled into easily recognizable geometrical shapes. All these segmentation approaches are generalized for urban structures and they do not produce meaningful results on complex architectural scenes. In the past, Manhattan world assumptions have been used for the reconstruction of architectural scenes from a single image ~\cite{koutsourakis2009single}.

Our work, however, uses a hybrid approach which includes model fitting, Manhattan world based projection sequences and a machine learning algorithm to generate a more accurate mix than the existing segmentation techniques.  We focus on 3D outdoor scenes with no restrictions towards planar surfaces or geometric buildings. Our model is a top-down hierarchical tree that iteratively divides the point cloud into smaller components using a greedy approach. The key idea behind our model is an energy function that automatically estimates the correctness of segmentation. Our energy function uses a neural network for classification thus resulting in a joint segmentation and detection framework. Here, we introduce a 3D convolution neural network, for the purpose of automatically classifying our 3D segments generated by the hirearichal tree. We use a novel voxelization technique which is applied on surface sampled points and we call this method HollowNets. The overall flow diagram of the proposed approach is shown in Fig.~\ref{fig:flow}.


\section{Challenges in LiDAR data}
\begin{figure}[tb!]
	\begin{center}\leavevmode \centerline{
		\begin{tabular}{@{}c@{}c@{}c@{}c@{}}    
				\includegraphics[width=.3\columnwidth]{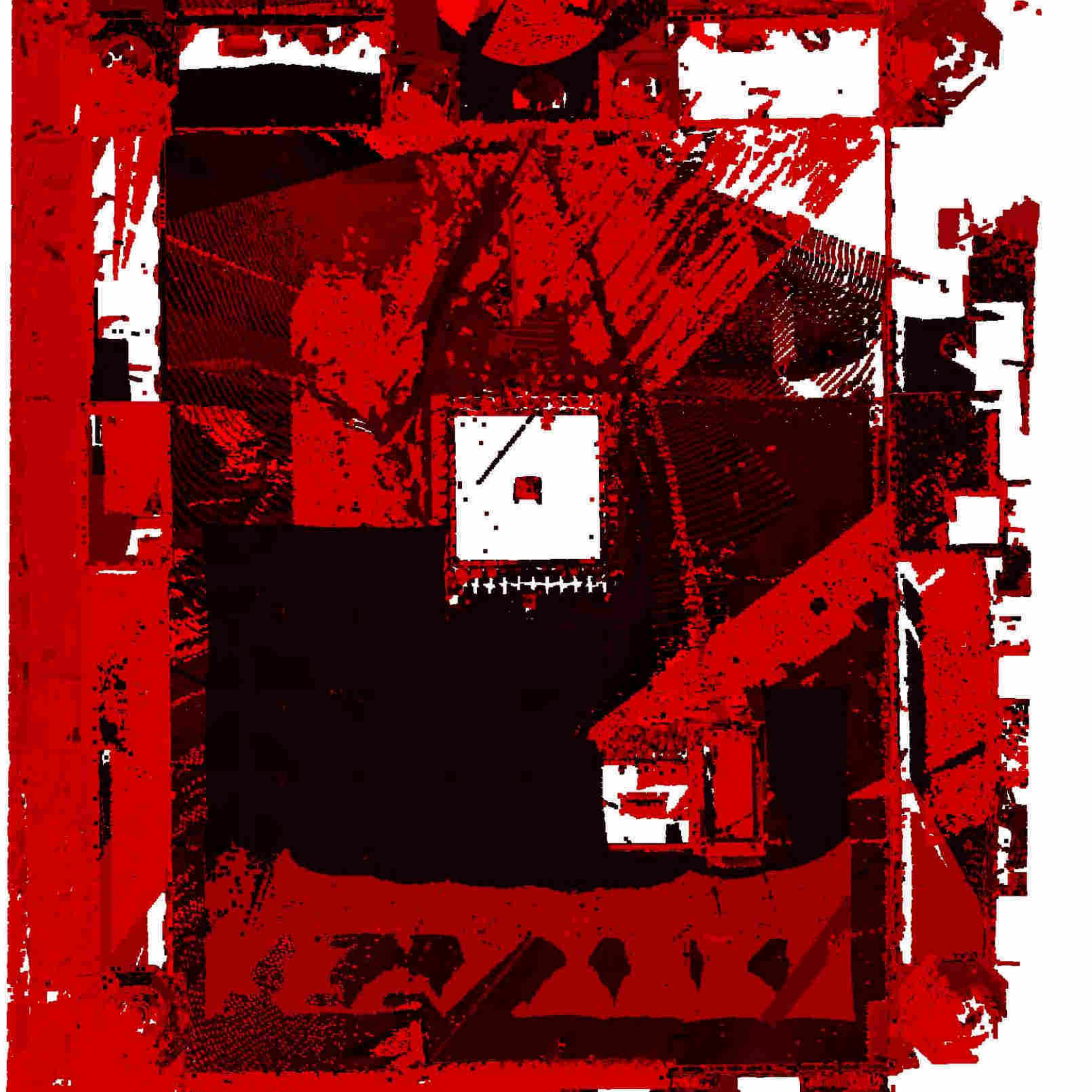} &	
        \includegraphics[width=.3\columnwidth]{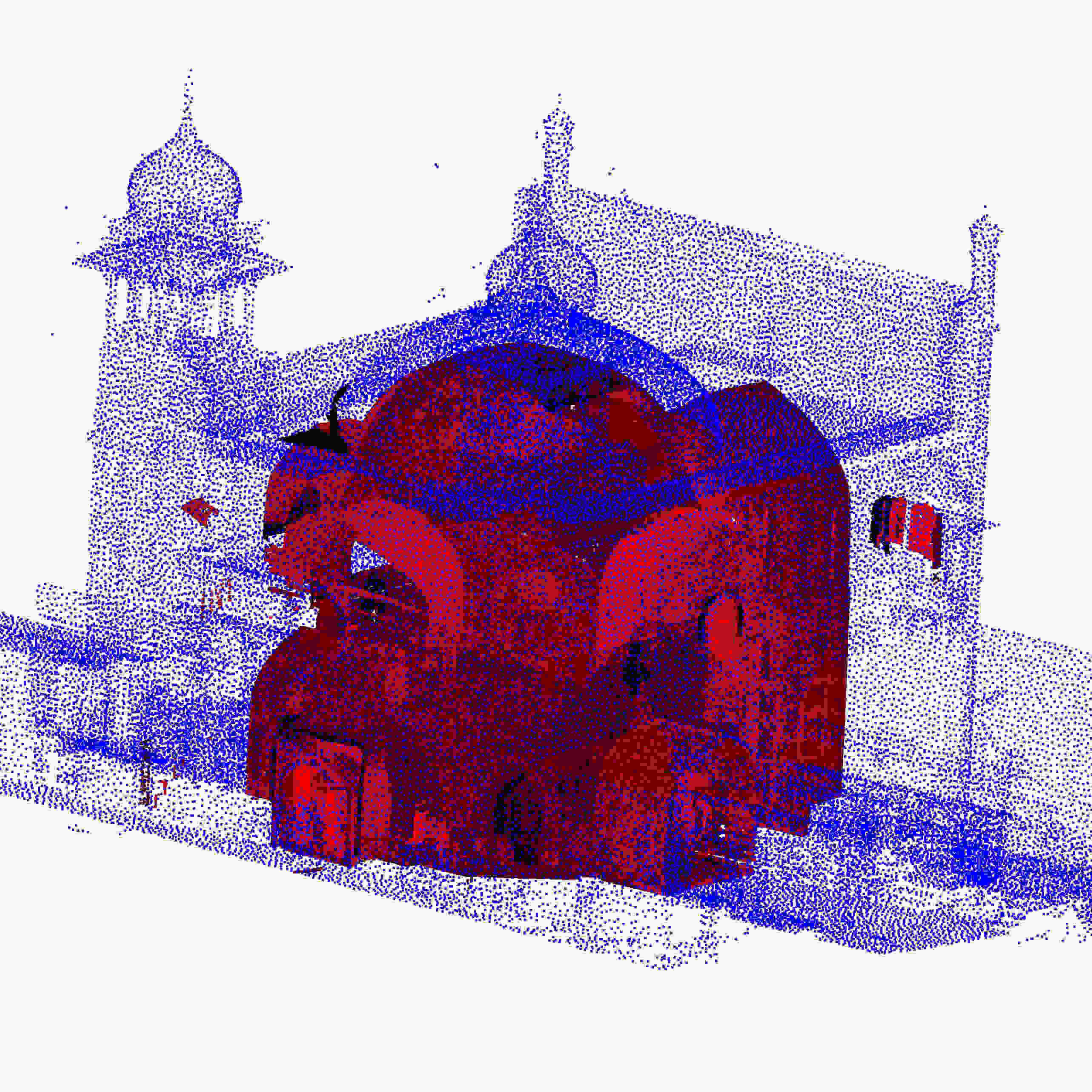} \\
                (a) & (b) \\
				\includegraphics[width=.3\columnwidth]{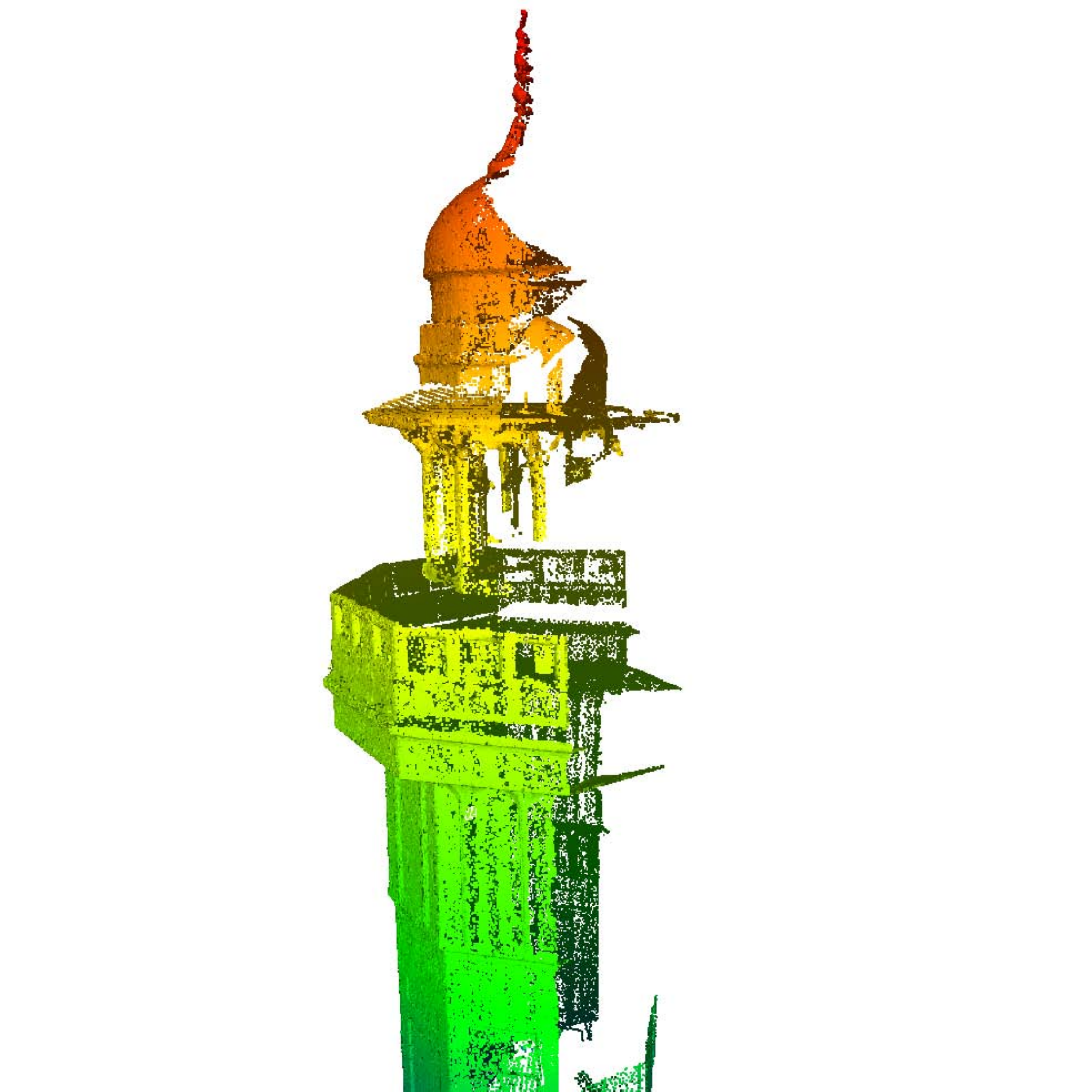} &				
				\includegraphics[width=.3\columnwidth]{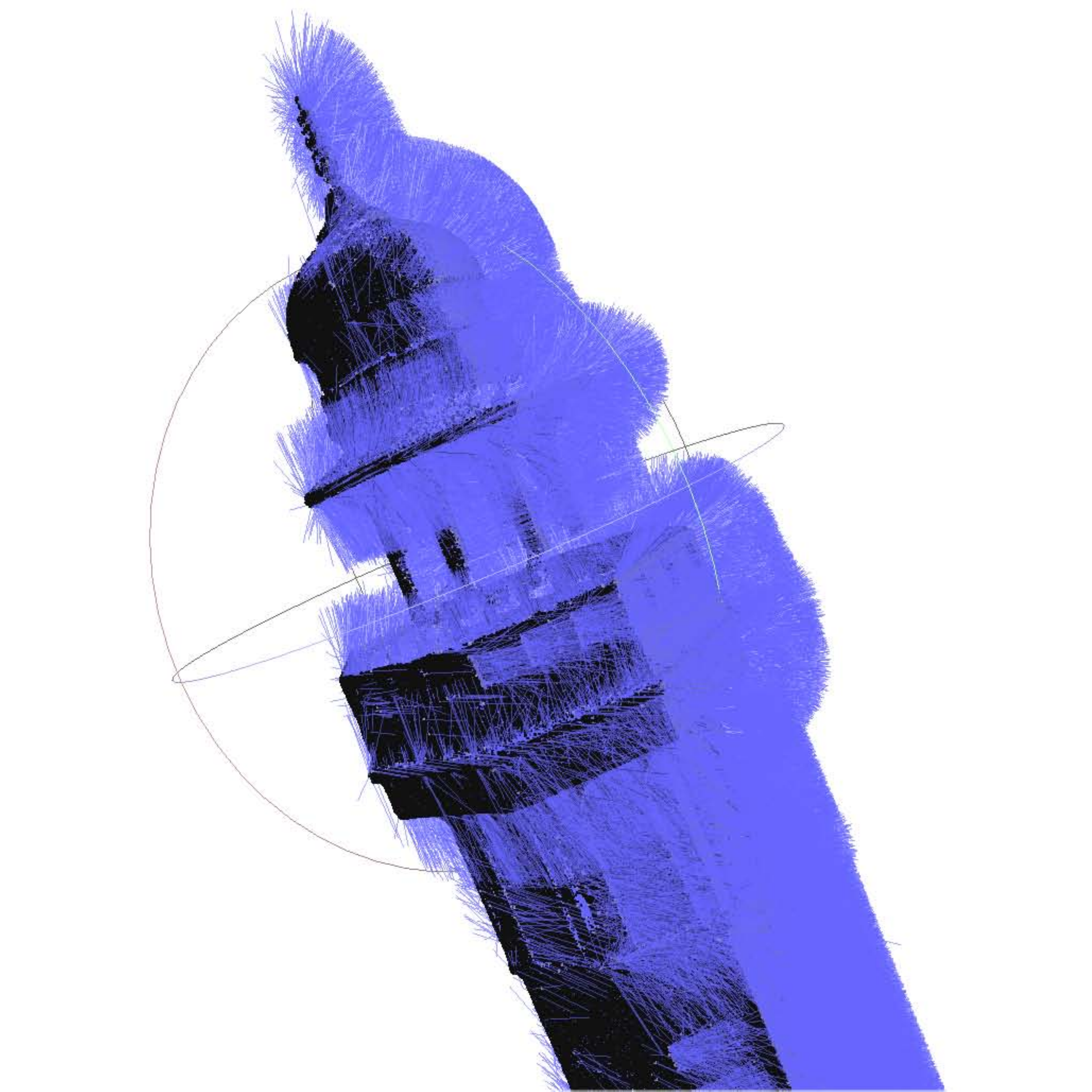}\\			
				(c) & (d)\\
		\end{tabular}
	} \caption{\label{fig:LiDARChallanges}
	LiDAR data challenges (a) Registration (b) Hollow Regions (c) Varying density and occlusion (d) Normals .}
    \end{center}
\end{figure}

Most of the previous work on point cloud, particularly using LiDAR, is either limited to indoor scenes~\cite{armeniCVPR16} or outdoor scene~\cite{Lin2013ToG} but not on both. Both static and moving platforms are used for laser scanning and are commonly referred to as Terrestrial Laser Scanning (TLS) and Mobile/Aerial Laser Scanning respectively. In this work, we have used TLS which allows us to obtain data that has both outdoor as well as indoor details of the architectural structure. Although LiDAR data is much more dense and accurate as compared to other sources such as Kinect, Tango and Photogrammetry, it poses several additional challenges that need to be addressed. Some of these challenges include data size, occlusion, hollow regions, normal computation, density and registration (see Fig.~\ref{fig:LiDARChallanges}).

\textbf{Registration}: Registration involves changing the position, orientation and scale of a point cloud with respect to a reference point cloud. This can either be done manually by defining corresponding 3D points or automatically using algorithms such as Iterative Closest Point (ICP)~\cite{ZhengyoIJCV1994}. Any error in estimating these parameters may result in erroneous data that may support multiple orientations in the overlapping regions. Since most segmentation algorithms use orientation as one of their major cues, they suffer from noisy registration. (see Fig~\ref{fig:LiDARChallanges}(a)). 


\textbf{Hollow regions}:
Outdoor LiDAR data such as aerial LiDAR only captures the outer surface of the structure, whereas indoor scans provide information about the inner surface. However, when a scanned data from indoor and outdoor scanning, performed using survey grade equipment such as TLS, is registered together this correctly results in two surfaces for each architectural component. It can be seen from Fig.~\ref{fig:LiDARChallanges}(b) that point cloud of exterior of the dome and its interior seen from the main prayer hall are very different. These variations are uncommon to benchmark training sets such as ShapeNet~\cite{WuCVPR2015} and other photogrammerty datasets. This results in significant recognition challenges when classifiers trained on benchmark 3DoR datasets are applied on such real datasets.

\textbf{Varying Density and Occlusion}:
The density of point clouds from Laser scanners depends upon its spacing and quality parameters. Furthermore, occlusions can also effect the point cloud density. Spacing indicates distance between two samples at 10m distance whereas quality is a measure of point precision. Since spacing is defined at 10m, this means objects near the scanner will be sampled at higher resolution and vice versa. Also, since a laser scanner can take up to a million points per second, the density of points also depends upon the number of scans in which that region was visible. This results in regions on the same surface having several times more points than other regions which are more occluded. Thus varying statics are observed for different regions on the same surface and hence degrade the segmentation performance (see Fig~\ref{fig:LiDARChallanges}(c)).

\textbf{Normals}:
Normals provide information about the orientation of the surface containing a fixed set of points. Thus they serve as an important cue for segmentation. Unlike photogrammetry, normal information is not available in the LiDAR data. For normal generation various techniques e.g. K-nearest neighbor, are used and any noise introduced by these techniques affect the resulting segmentation. 
To address some of these challenges, we propose a hierarchical tree model which is discussed next.
\section{Hierarchical Tree Model}
3D scene understanding is commonly addressed as a detection problem or segmentation problem. In detection, exhaustive scanning of an entire scene is computationally unfeasible for large-scale scenes while space based partitioning such as Octree is prone to splitting objects. We, instead, perform segmentation by recursively partitioning the points in a subspace spanned by the projection of the points into a 1D signal. 

The input to our algorithm is coarse segments obtained using RANSAC based primitive fitting~\cite{schnabel-2007-efficient} and spatial clustering. Our partitioning space is inspired by the work of Bajwa et al.~\cite{BajwaCGI2016} and Hassaan~\cite{HassaanPG2017} that employs the Manhattan world assumption and projects the 3D data into one or more orthogonal planes. However, their projections do not directly provide any segmentation. They proposed a semi-interactive approach in which the object class information was manually provided for each of the point cloud segments. The correct projection was then applied based on heuristics for each object category. 

\begin{figure}[b]
\centering
\includegraphics[width=.99\linewidth, trim={0 500 0 50},clip]{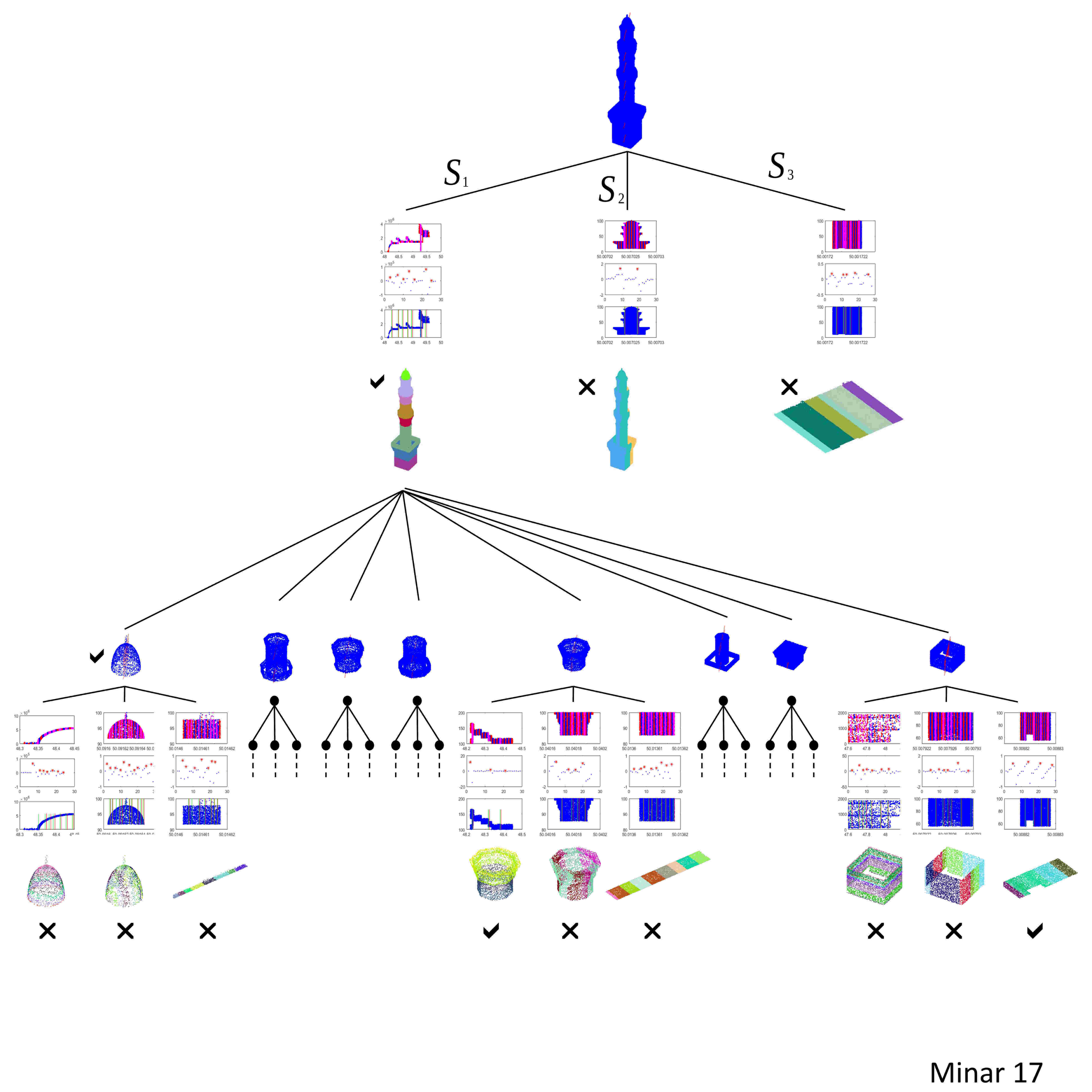}
\caption{\label{fig:htt}
	Example showing generation of tree when segmentation is being applied on a minar. The Minar is segmented using all three projection sequences and then the $8$ segments produced by $S_1$ are selected based on energy. Each of them are then re-segmented.}
\end{figure}

On the other hand, we have used three generalized projection sequences each of which results in a sub-segmentation of the point cloud data. Hierarchical organization of structural elements at different scales and locations are commonly seen in man-made structures~\cite{pauly2008discovering}. Contrary to Lin's~\cite{Lin2013ToG} bottom-up hierarchical tree of planar patches, our's is a top-down model of hierarchical segmentation. We recursively build a tree $G\left \{ V,E \right \}$ where the edge $E$ is one of the projection sequences and the nodes $V$ are the obtained sub-segments. The weight of each of these edges is computed using an objective function $\xi(v_{i,j}^n,v_{i-1,k}^{m})$ defined as

\begin{equation}
\xi(v_{i,j}^n,v_{i-1,k}^{m}) = \mathbf{\omega}^T\mathbf{\epsilon},
\label{eq:energy}
\end{equation}

where $\mathbf{\omega}=\{\omega_1, \cdots, \omega_5\}$ are the weights for each of the five energy terms $\mathbf{\epsilon}=\{e_1\cdots,e_5\}$. These weights are obtained using simple linear regression. The node $v_{i,j}^n$ is the $j^{th}$ segment of the $i^{th}$ iteration (tree depth) obtained via $n^{th}$ projection sequence such that $v_{i,j}^n \subset v_{i-1,k}^{m}$ and $v_{i-1,k}^{m}$ is the parent node of $v_{i}$. Also $v_{i-1,k}^{m} = \{\cup v_{i,j}^n\}$ where, for all segments $s$, $j \in \{1,2,\dots, s\},~ n \in \{1,2,3\}$. The set $V_{i}^{n}$ contains all segments of $v_{i-1,k}^{m}$ obtained via $n^{th}$ projection sequence. Each of the projection sequences is discussed next and energy terms are discussed in Sec.~\ref{sec:objfunc}.

\subsection{Projection sequences}
We receive coarse segments from the Schnabel's algorithm~\cite{schnabel-2007-efficient}, which are then clustered based on spatial proximity. In the next step, the tree methodology is applied on each coarse segments. As shown in the Fig.~\ref{fig:htt}, on each segment, three different projection sequences are applied. Peak finding $\rho(.)$ is then applied on the obtained low-dimensional signal to obtain the segments. The input point cloud is converted into low-dimensional signal using four projections namely vertical ($\upsilon_1 \& \upsilon_2$), horizontal ($h$), circular profile ($p$) and circular un-warping ($u$). 

The projections $\upsilon_1$ and $h$ involves eliminating one of the dimensions. Circular and n-gonal RANSAC $\psi(.)$ are then applied on the projected data to recover the location, position and orientation. Since peak finding $\rho(.)$ can return multiple peaks belonging to the same segmentation boundary, operation $\upsilon_2$ is first applied to convert the data into a limited bin histogram. The optimal number of bins differs for each point cloud and is computed by counting the zero-crossings in the derivative signal. Finally, peak finding $\rho(.)$ is performed to obtain the segments.



\begin{figure}[tb!]
\centering
\includegraphics[width=.99\linewidth,]{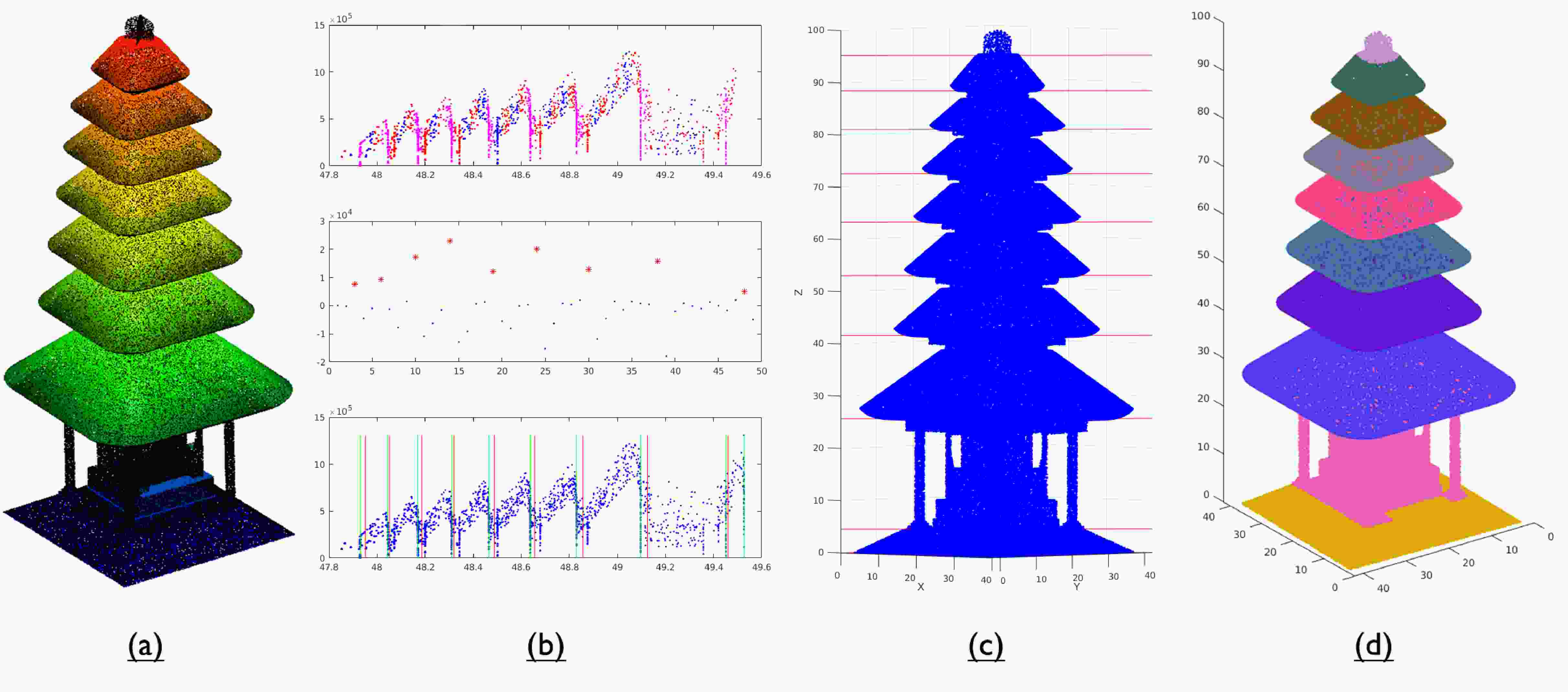}
\caption{\label{fig:temple_detail}
Example showing detailed modeling via profile curve. (a) Point cloud. (b) Peak finding on profile curve. (c) Peak locations on point cloud. (d) Detailed segmentation.}
\end{figure}

\textbf{Sequence $S_1$}: $V_{i}^{1} = \rho(\upsilon_2(p(\psi(\upsilon_1(v_{i-1,k}^{m})))))$\\
In order to generate a profile curve, a transport curve is needed which is obtained by applying a RANSAC algorithm on a vertically projected point cloud. We then generate the histogram $\upsilon_2$ of the profile curve, the peaks of the histogram are then refined within the corresponding bin to find the actual peak. The peaks facilitate the segmentation of the input point cloud. Figure~\ref{fig:temple_detail} demonstrates the working of sequence $S_1$.

\textbf{Sequence $S_2$}: $V_{i}^{2} = \rho(\upsilon_2(h(v_{i-1,k}^{m})))$\\
This branch first performs a horizontal projection (the y-axis points of the point cloud are equated to zero) and then the histogram $\upsilon_2$ is used to find the peaks.

\textbf{Sequence $S_3$}: $V_{i}^{3} = \rho(\upsilon_2(u(\psi(\upsilon_1(v_{i-1,k}^{m})))))$\\
This segmentation sequence is similar to Sequence 1 except that instead of finding the profile curve, the data is unwarped using the transport curve.

All three projection sequences are applied recursively on each of the obtained sub-segment. Each projection sequence generates a set of sub-segments. This is a greedy approach and only the set of sub-segments having the highest weighting edge is selected for further segmentation. The remaining two segment sets are discarded, if the energy of all the segment sets is below a specified threshold the node $v_{i-1,k}^{m}$ is declared as a leaf node. Once the required recursion depth is reached, all the segments $v_{i,j}^{n}$ in the set having the highest energy are added as leaf nodes.

\subsection{Objective function}
\label{sec:objfunc}
Ideally, only the sequence that would yield correct sub-segments should be applied on each segment. However, it is not possible to identify the correct next projection a priori. Since the goal of projection sequence is to simplify the data thus facilitating the segmentation, an incorrect projection sequence, e.g. a circular unwarp of an archway or bridge, would have adverse effects on the data. We estimate the information content and goodness of the obtained segment set based on $5$ criteria. Hence, the energy function $\xi(.)$ that we use contains $5$ terms based on these criterion:

\begin{itemize}

\item  Correct segmentation will have more or less uniform distribution of points among segments. In order to avoid skewed distribution of points among segments, the normalized deviation between the segment population $\epsilon_1$ is computed and is defined as:
\begin{equation}
	\epsilon_1 = 1 - \frac{\sigma([N_1,N_2,\dots, N_s])}{\sigma([1, N_p-1])}.
	\label{eq:E3}
\end{equation}

\item Correct segments can only be obtained from a segment having considerably large number of points. Hence, the parent node population score $\epsilon_2$ is defined as the ratio between number of points $N_p$ and the set threshold on $N_{min}$. This energy is maximum in case $N_p \geq N_{min}$ and is defined as
\begin{equation}
	\epsilon_2 = \frac{\min (N_p, N_{min})}{N_{min}}.
	\label{eq:E2}
\end{equation}

\item Correct segmentation will neither produce a large number of segments nor will it give a single segment only. Segmentation resulting in only a single segment is penalized using a Gaussian distribution having $(\mu, \sigma) = (1,1)$ and the resulting energy $\epsilon_3$ is defined as
\begin{equation}
	\epsilon_3 = 1 - \sim \mathcal{N} (s | \mu,\sigma^2).
	\label{eq:E4}
\end{equation}

\item Some projection sequences will always be inapplicable on certain object categories. To introduce a semantic relationship between nodes, $\epsilon_4$ incorporates a prior probability of observing a certain projection sequence given the class information of the initial segment and is defined as
\begin{equation}
	\epsilon_4 = W(r_{ID}(v_{i-1,k}^{m}), seq. ID),
	\label{eq:E5}
\end{equation}
where $W$ is a $K\times 3$ matrix containing prior probabilities of each path for each object class. 

\item Each sub-segment of a correctly segmented point cloud will have a higher recognition score.
The last energy term $\epsilon_5$, is thus based on the classification score $r_{scr}$ of each segment and is defined as:

\begin{equation}
	\epsilon_5 = \frac{1}{s}\sum{r_{scr}(v_{i,j}^{n})},~j \in \{1,2,\dots, s\}.
	\label{eq:E1}
\end{equation}
\end{itemize}

The applied segmentation process is shown in Algorithm~\ref{algo:treealgo}.

\begin{algorithm}
\small
\caption{Generate Hierarchical Tree}
\label{ngonAlgo}
\begin{algorithmic}[0]
\begin{spacing}{1.1}
\Procedure{GenerateHierarchicalTree}{$p,d$}
\LeftComment{$p$ is a $N\times 3$ matrix of 3D points}
\LeftComment{$d$ is a maximum allowed recursion depth}
\State $G \leftarrow p$ \Comment{add node}
\State $V_{i}^{1} = \rho\Bigg(\upsilon_2\bigg(p\Big(\psi\big(\upsilon_1(p)\big)\Big)\bigg)\Bigg)$
\State $V_{i}^{2} = \rho\Big(\upsilon_2\big(h(p)\big)\Big)$
\State $V_{i}^{3} = \rho\Bigg(\upsilon_2\bigg(u\Big(\psi\big(\upsilon_1(p)\big)\Big)\bigg)\Bigg)$
 
\For {$each V_{i}^{n}$}
	\State \textbf{compute} energy $\xi^n(.)$ using eq.~\ref{eq:energy}
\EndFor
	
\State $\hat{\xi} = \max_{n}\xi^n :=\{n | \forall \xi^m \leq \xi^n\}$
	\If {$\hat{\xi} < threshold$}
		\State \textbf{return;} \Comment {Invalid segmentation}
	\Else
	  \If	{$d > zero$}
			\For {each $v_{i,j}^{n} ~in ~V_{i}^{n}$ }
				\State {$GenerateHierarchicalTree(v_{i,j}^{n},d-1)$}
			\EndFor
		\EndIf
	\EndIf
	
\EndProcedure
\end{spacing}
\end{algorithmic}
\label{algo:treealgo}
\end{algorithm}


Where the set $V_{i}^{n}$ contains all segments of $v_{i-1,k}^{m}$ obtained via $n^{th}$ projection sequence. The classification is performed using a proposed \textit{HollowNet} CNN which is discussed in the next section. 
\subsection{Deep Learning: HollowNet}
Unlike 2D images, a point cloud contains variable amount of data and has irregular dimensions in 3D space. Thus, in order to use it in machine learning algorithms we sub-sample data into a regular grid. Seminal work in 3D object recognition such as VoxNet\cite{MaturanaIROS2015} and ShapeNet\cite{WuCVPR2015} uses a volumetric representation of objects.

\textbf{Voxel Representation}: Since our point cloud is a surface data instead of volumetric data, we use a simplistic approach towards the representation of 3D point cloud data in terms of regular voxel grid. We scale each point cloud object to our voxel size and fit the point cloud inside our 3D cube thus mapping each $(x,y,z)$ point location to $(i,j,k)$ index of 3D regular grid. We keep track of each point to index mapping in a separate file so that we can regenerate the actual point cloud using its voxel representation. Laser scanners provide surface representation of surrounding objects only, thus, we call our voxel representation as Hollow voxel. Our training dataset consists of 7 classes. Our  including unique primitives not found in 3D outdoor scene datasets. This data was collected from various free online resources.
\vspace{-5px}

\begin{figure}[b]
\centering
\includegraphics[width=.99\linewidth]{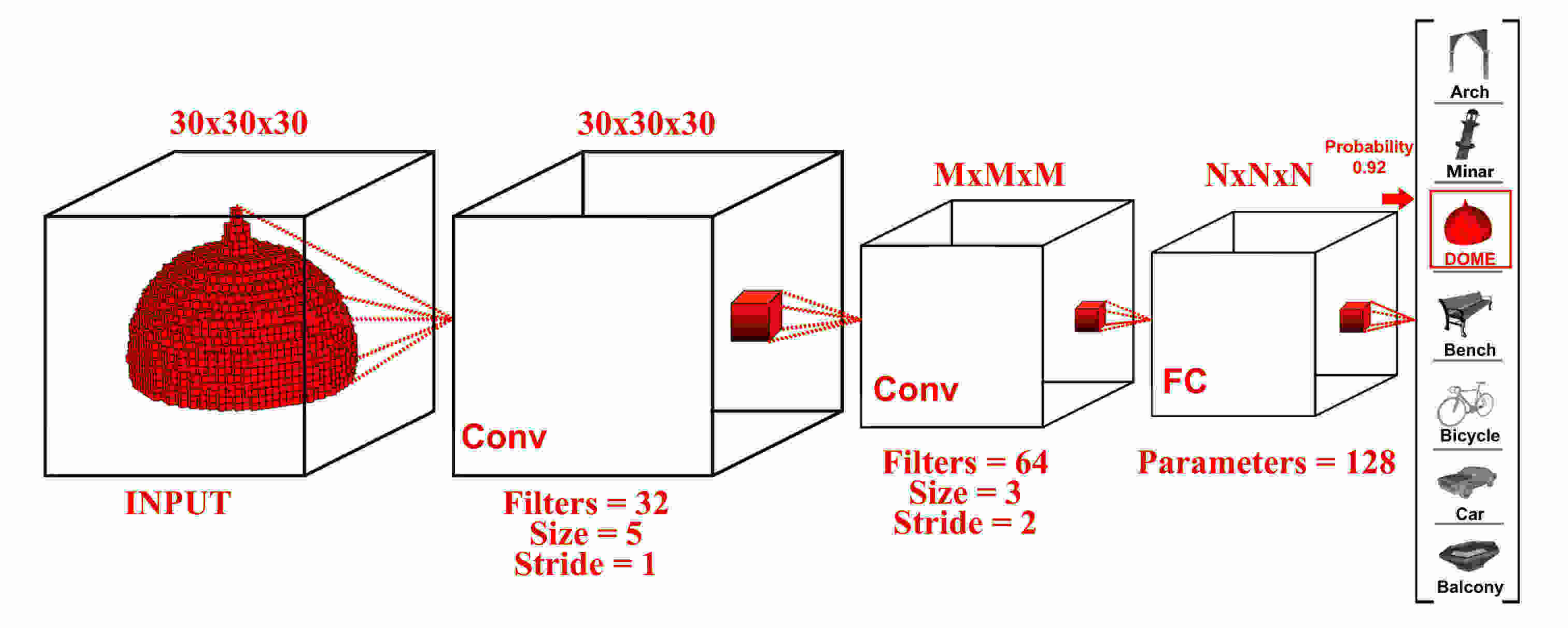}
\caption{\label{fig:layers}
	Layered Architecture used to train HollowNet.}
\end{figure}

\begin{figure}[t]
\centering
\includegraphics[width=.4\columnwidth] {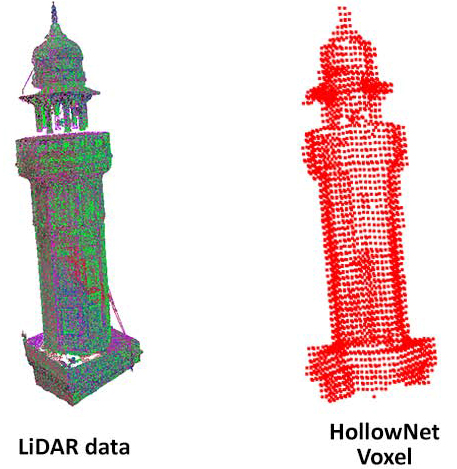}
\caption{\label{fig:test}
 Sample illustration of hollow voxel generated for surface based mesh,  a minaret. 
}
\end{figure}

\textbf{CNN}: The input to our neural network is a voxel volume of size  $L \times W \times H$, where $L$ is the length, $W$ is the width and $H$ is the height of voxel data. Currently we are using the resolution of $30 \times 30 \times 30$. The prediction problem requires producing a target output of size \textit{K} which is the number of classes we have used to train our network. The Neural network outputs a vector of length equal to the length of number of classes to show the probability of each class. For our problem we have chosen $K = 7$ each having $300$ and $50$ training and testing samples, respectively collected from 3D Warehouse and ModelNet10 ~\cite{WuCVPR2015}. We obtained an accuracy of $99.03\%$ on $1000$ EPOCHS at a learning rate of $\eta=0.001$.


We thus implement a surface-based \textit{hollow-voxel} representation in which each 3D point is first projected to a canonical view volume. Each $(x,y,z)$ location in pointcloud is then mapped to $(i,j,k)$ index of 3D regular grid (see Fig.~\ref{fig:test}). 

Each cross-section of this representation can be considered as an output of convolution by an edge filter. This makes our representation inherently suitable for commonly used deep convolution neural networks (CNN). Furthermore, instead of using binary voxels, we scale values between $[-1, 5]$ thus allowing the neural network to learn more from positive integer values.


\textbf{Layered Architecture}: Contrary to existing volumetric voxel-based CNN, learning on hollow-voxels can be performed on a much simpler network architecture due to their inherent gradient like representation. Thus, the model that we use(see Fig.~\ref{fig:layers}) consists of $2$ convolutional layers, each performing 3D convolution with a bank of $5^3$ dimensional filters, starting with $32$ filters and then doubling in the next layer. Strided convolution in the second layer performs the functionality of pooling layer. Data layer takes a $4$ dimensional input volume, in which $3$ dimensions are spatial and the fourth represents the batch size.

All convolutional layers use leaky rectified linear units (Leaky ReLUs) as a non-linearity. This neural network consists of $2$ fully connected layers. Each fully connected layer consists of $n$ neurons where the output of each neuron is a linear combination of previous input neurons of the neural network. For the second fully connected layer, the number of neurons is equal to number of desired output classes $K$. For our experiments we have used K$=7$.

\begin{table}[t]
\centering
\caption{Confusion Matrix of Neural Network Input Classes}
\label{my-label}
\begin{tabular}{|l|l|l|l|l|l|l|}
\hline
        & Bench & Car & Bicycle & Arch & Dome & Minar \\ \hline
Bench   & 245   & 00  & 0       & 0       & 0    & 0     \\ \hline
Car     & 6     & 300 & 2       & 0       & 0    & 0     \\ \hline
Bicycle & 1     & 0   & 195     & 0       & 0    & 0     \\ \hline
Arch  & 0     & 0   & 0       & 130     & 0    & 0     \\ \hline
Dome    & 1     & 0   & 2       & 0       & 120  & 0     \\ \hline
Minar   & 3     & 1   & 0       & 0       & 0    & 130   \\ \hline
\end{tabular}
\end{table}


\section{Results}
\subsection{Experimental Setup}
We evaluated our algorithm both on synthetic as well as real data. The synthetic scenes  consisted of two buildings, namely Temple and Roman building, which are scaled-down real world scenarios having $0.5$ to $2.5$ million points. The real data was obtained from Leica Scan Station P20 Terrestrial Laser Scanner. Three sites were scanned using P20 namely Derawar Fort, Masjid Wazir Khan and Masjid Khudabad. The combined data of these three sites is around $1.36$ billion points. The details of these sites and obtained results are shown in Table.~\ref{tab:compdata}. The data will be made public via our website along with the paper.
\begin{table*}[tb]

\centering
\caption{\label{tab:compdata}Results showing comparison of automatically generated (AG) primitives with ground truth (GT) and accuracy (Acc\%).}
%
\begin{tabular}{|c|c|c|c|c|c|c|c|c|c|c|c|}
\hline
\textbf{Site} & \textbf{Dimensions}           & \textbf{Points} & \multicolumn{3}{c|}{\textbf{Arches}} & \multicolumn{3}{c|}{\textbf{Domes/Chattri}} & \multicolumn{3}{c|}{\textbf{Minarets/Pillars}} \\ \hline
              & $L\times W\times H~m^3$ & in Bn           & GT        & AG        & Acc        & GT        & AG        & Acc       & GT         & AG         & Acc        \\ \hline
Masjid Wazir Khan           & $91\times 53\times 33$                      & $0.288$           & $46$        & $19$        & $41.30$        & $12$        & $12$        & $100$         & $6$          & $6$          & $100$          \\ \hline
Masjid Khuadabad           & $60\times 36\times 16$                      & $0.548$           & $12$        & $7$         & $58.33$        & $21$        & $19$        & $90.4$        & $0$         & $0$          & $NA$           \\ \hline
Derawar Fort           & $1500\times 1300\times 30$                      & $0.43$           & $0$        & $0$         & $NA$        & $0$        & $0$        & $NA$        & $38$         & $38$          & $100$           \\ \hline
Roman building           & $20\times 36\times 14$                      & $0.02548$           & $79$        & $77$         & $97.44$        & $0$        & $0$        & $NA$        & $0$         & $0$          & $NA$           \\ \hline
Temple           & $10\times 16\times 44$                      & $0.0154$           & $0$        & $0$         & $NA$        & $7$        & $7$        & $100$        & $4$         & $0$          & $0$           \\ \hline

\end{tabular}

\end{table*}

\begin{figure}[h!]
\centering
\includegraphics[width=.7\columnwidth] {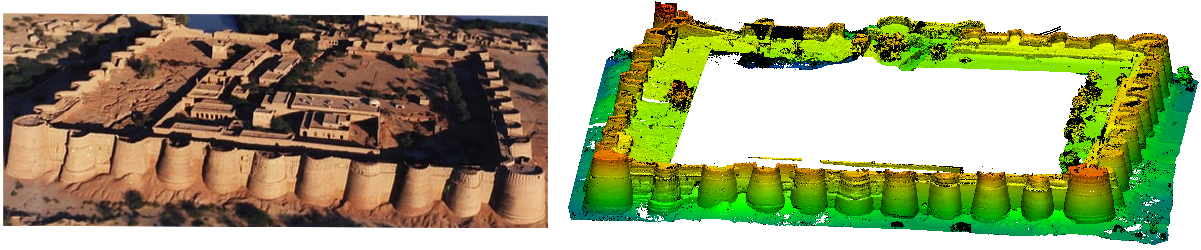}
\includegraphics[width=.7\columnwidth] {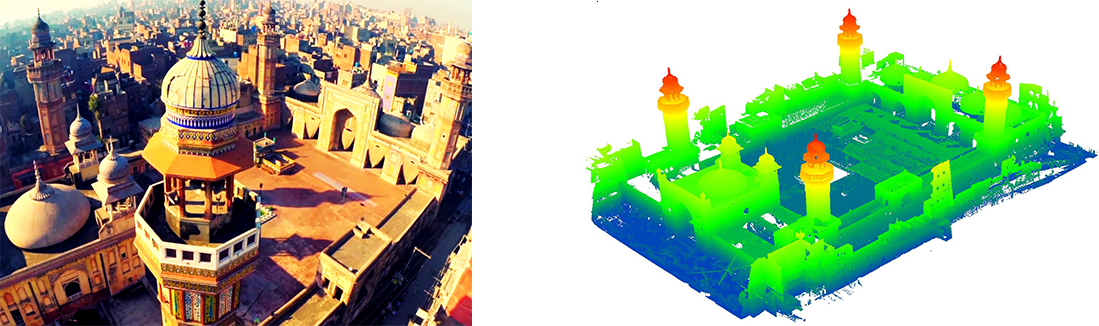}
\includegraphics[width=.7\columnwidth] {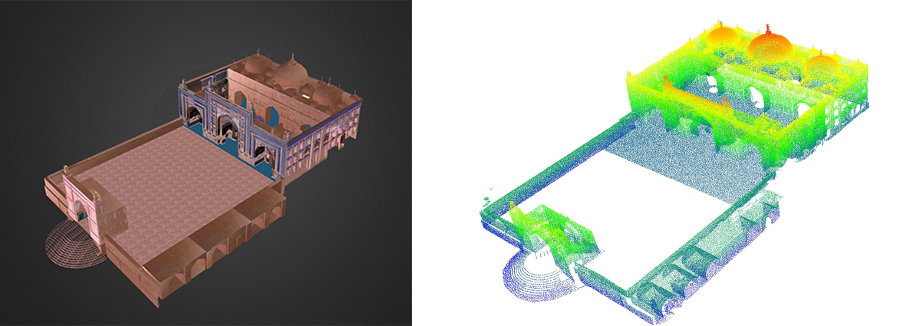}

\caption{\label{fig:derawar_teaser}
LiDAR data set. (Row 1) Derawar fort. (Row 2) Masjid Wazir Khan. (Row 3) Masjid Khudabad.}
\end{figure}


\subsection{Experiment: Synthetic Point Cloud}
\begin{figure*}[th!]
	\begin{center}\leavevmode \centerline{
		\begin{tabular}{@{}c@{}c@{}c@{}c@{}}
				\includegraphics[width=.5\columnwidth]{images/coll_original} &	
        		\includegraphics[width=.5\columnwidth]{images/colloseum_schnabel3} &
				\includegraphics[width=.5\columnwidth]{images/colloseum_cluster6} &
				\includegraphics[width=.5\columnwidth]{images/colloseum_output3}\\
				\includegraphics[width=.5\columnwidth]{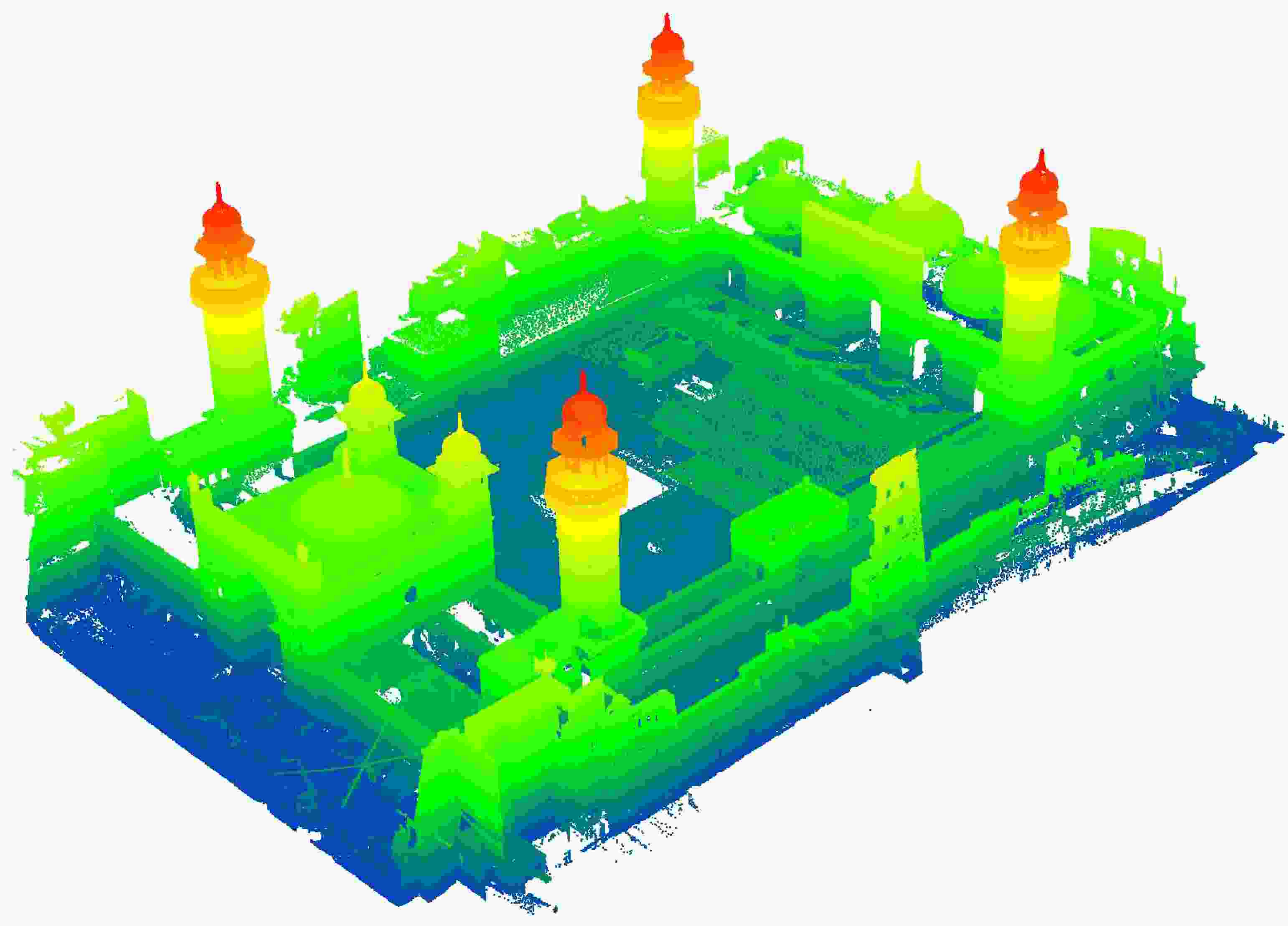} &	
				\includegraphics[width=.5\columnwidth]{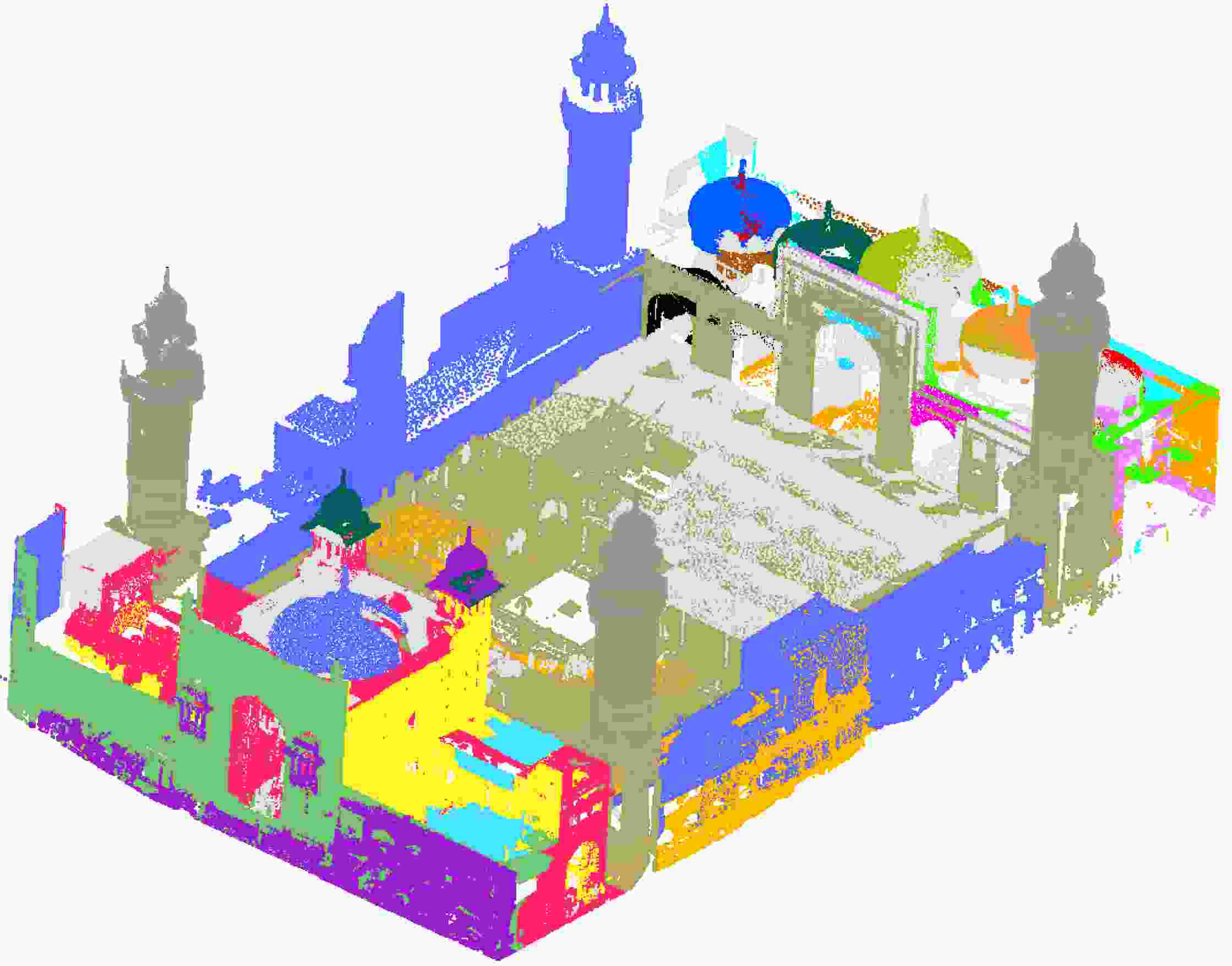} &	
        		\includegraphics[width=.5\columnwidth]{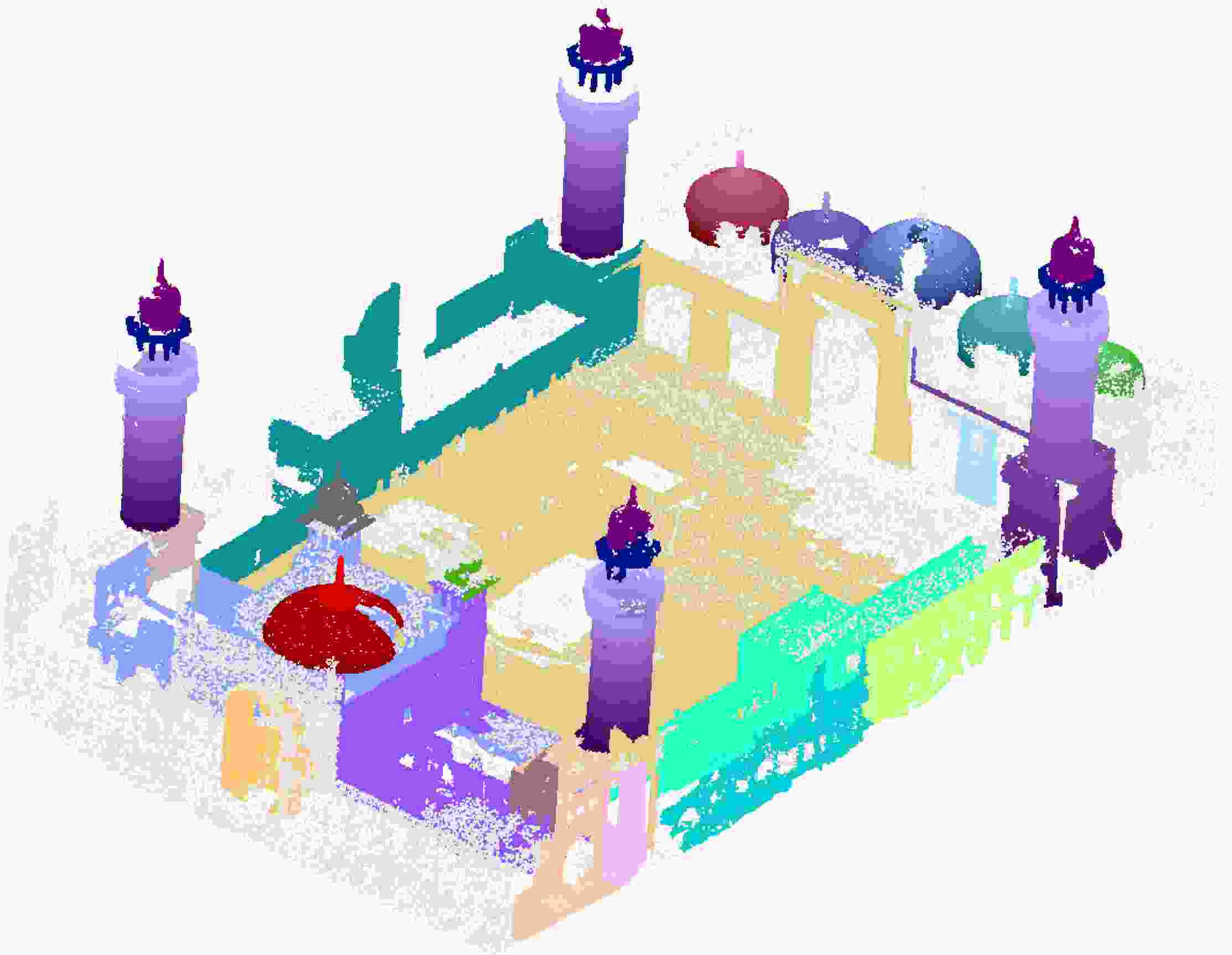} &
				\includegraphics[width=.5\columnwidth]{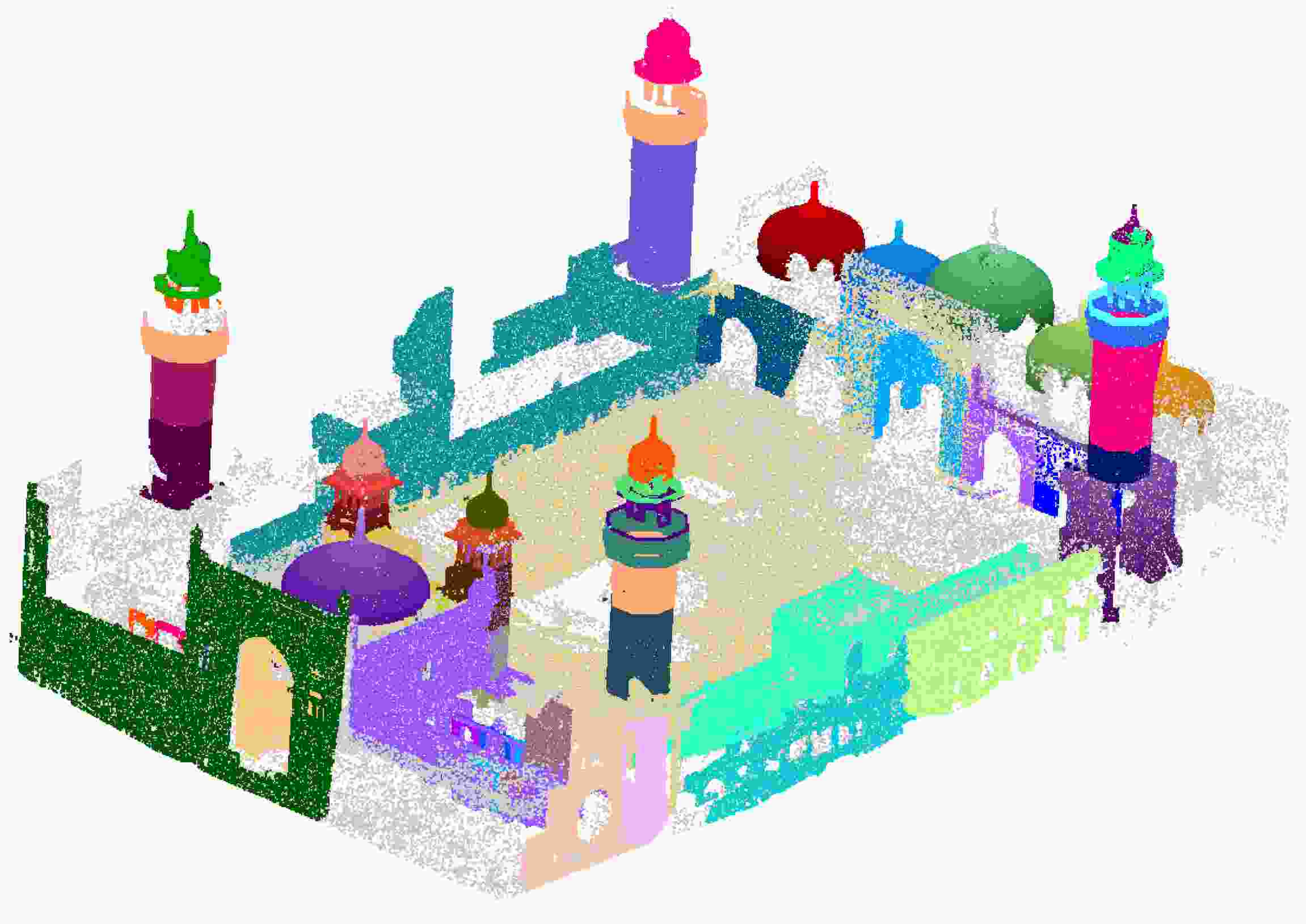} \\
				\includegraphics[width=.5\columnwidth]{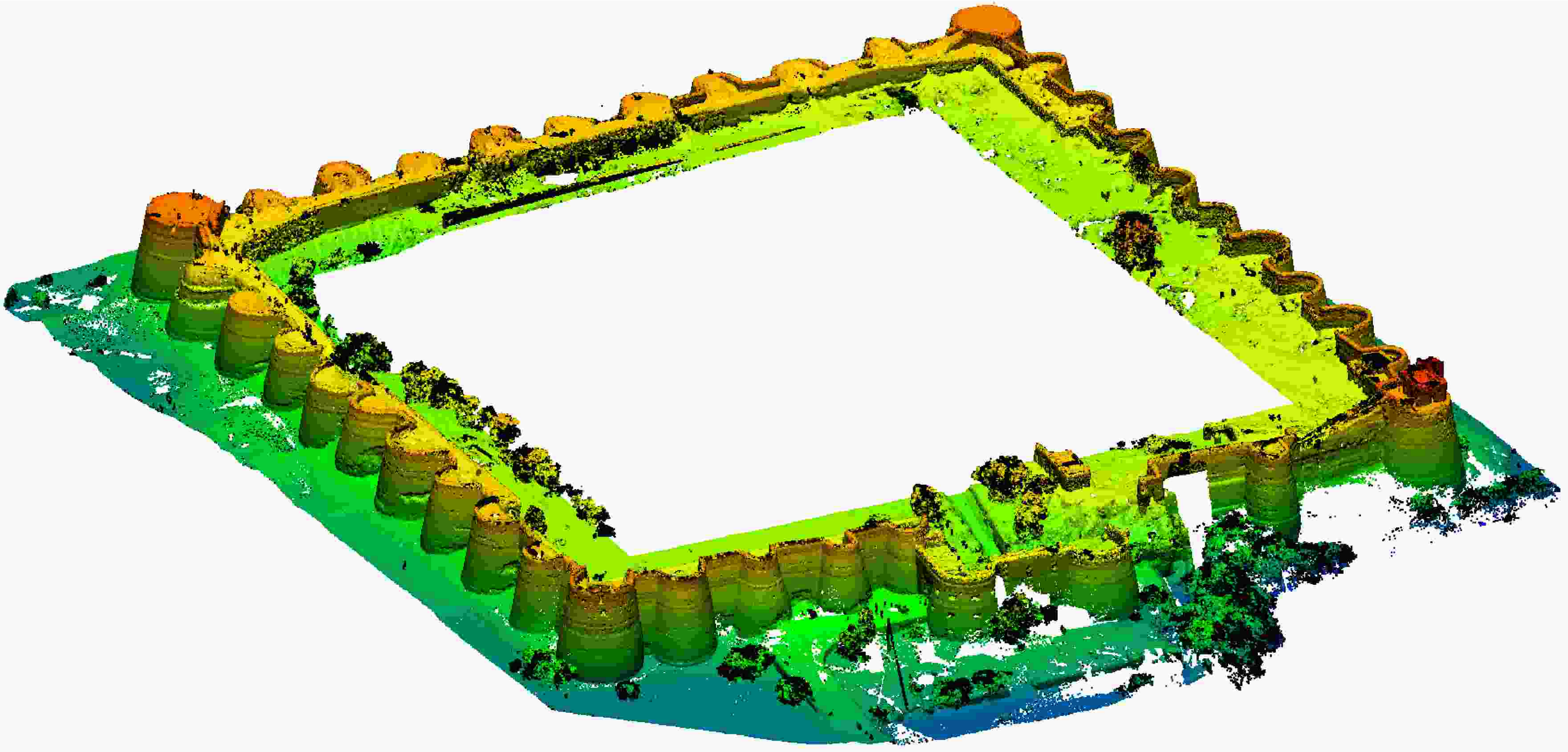} &	
        		\includegraphics[width=.5\columnwidth]{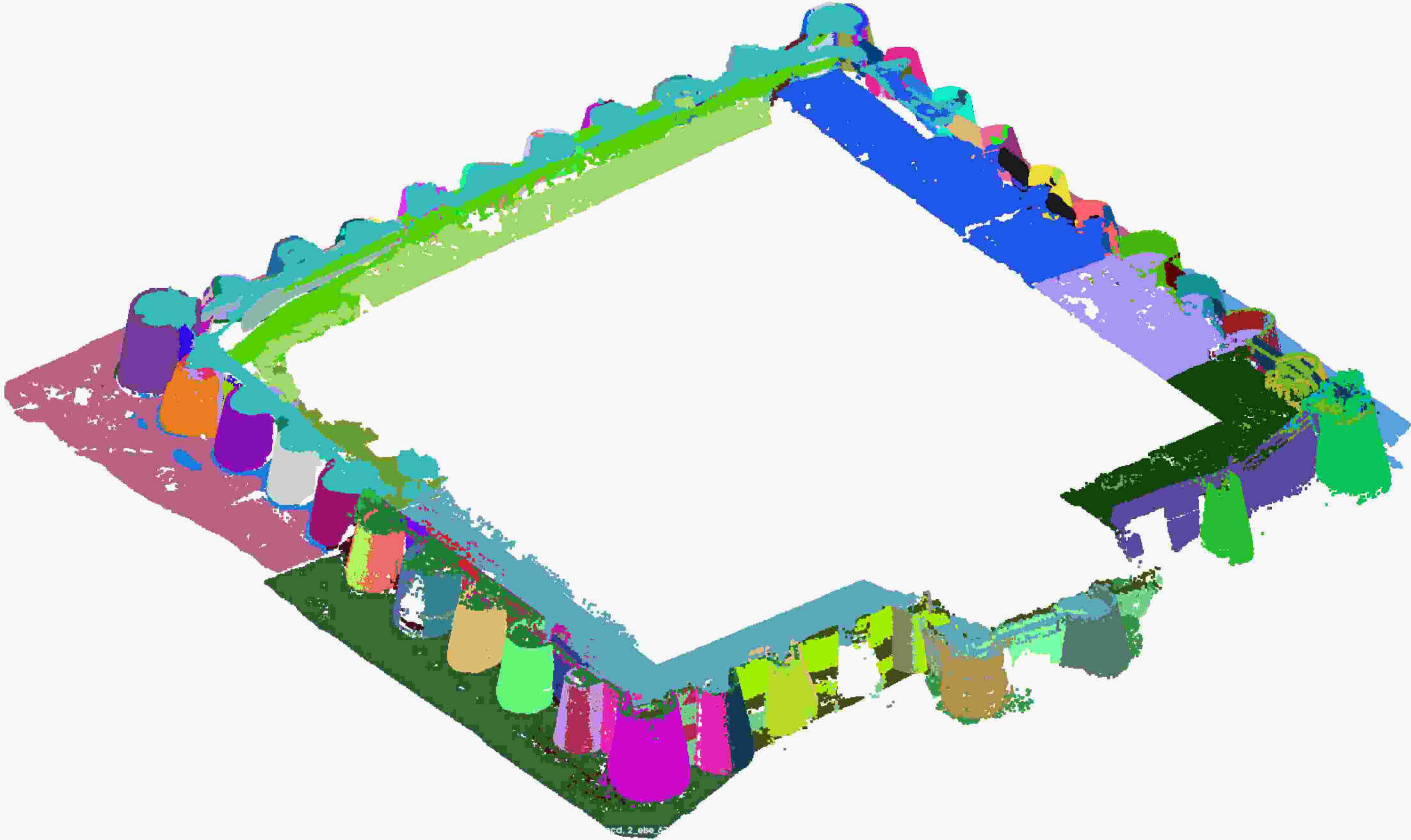} &
				\includegraphics[width=.5\columnwidth]{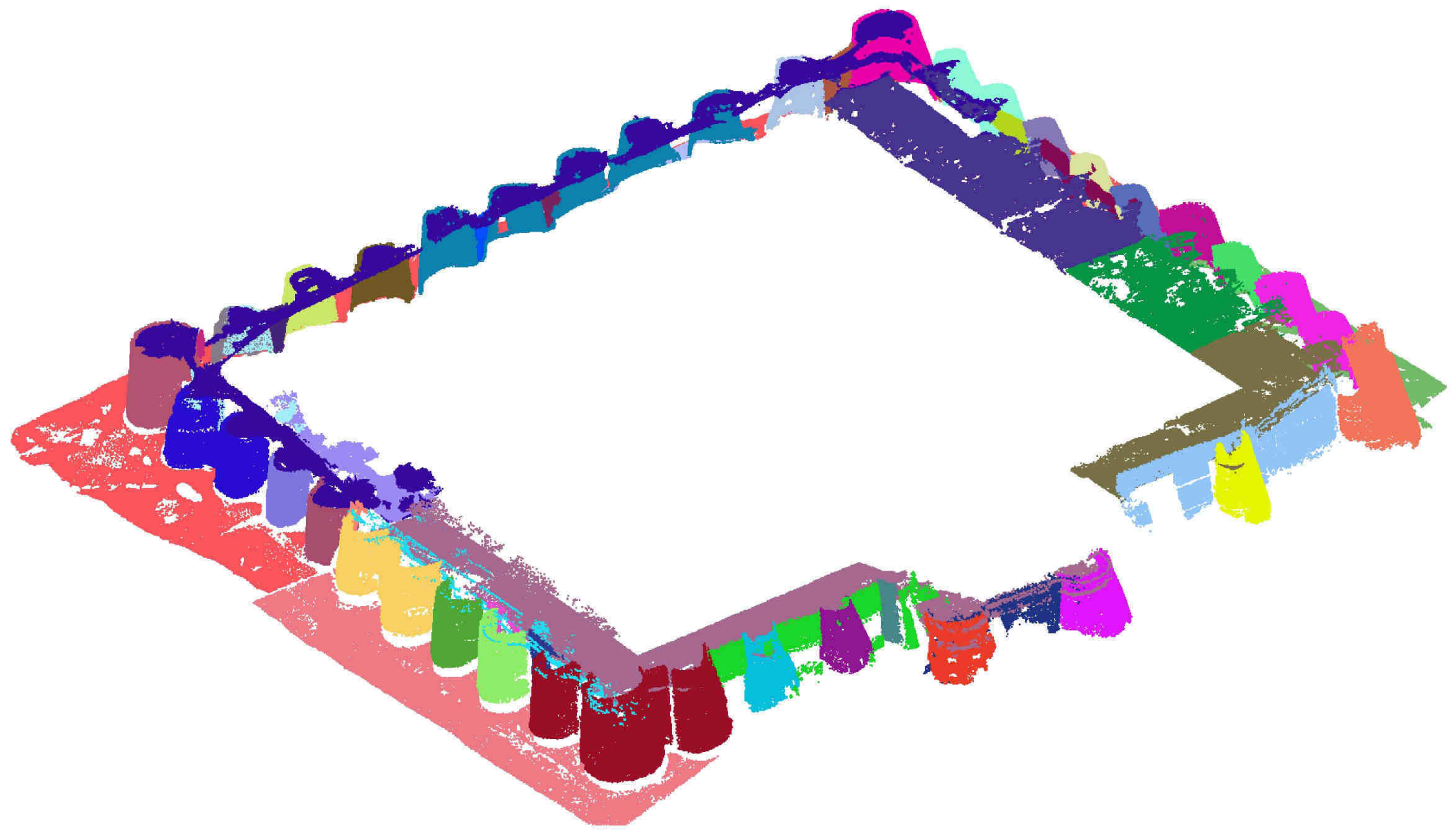} &
				\includegraphics[width=.5\columnwidth]{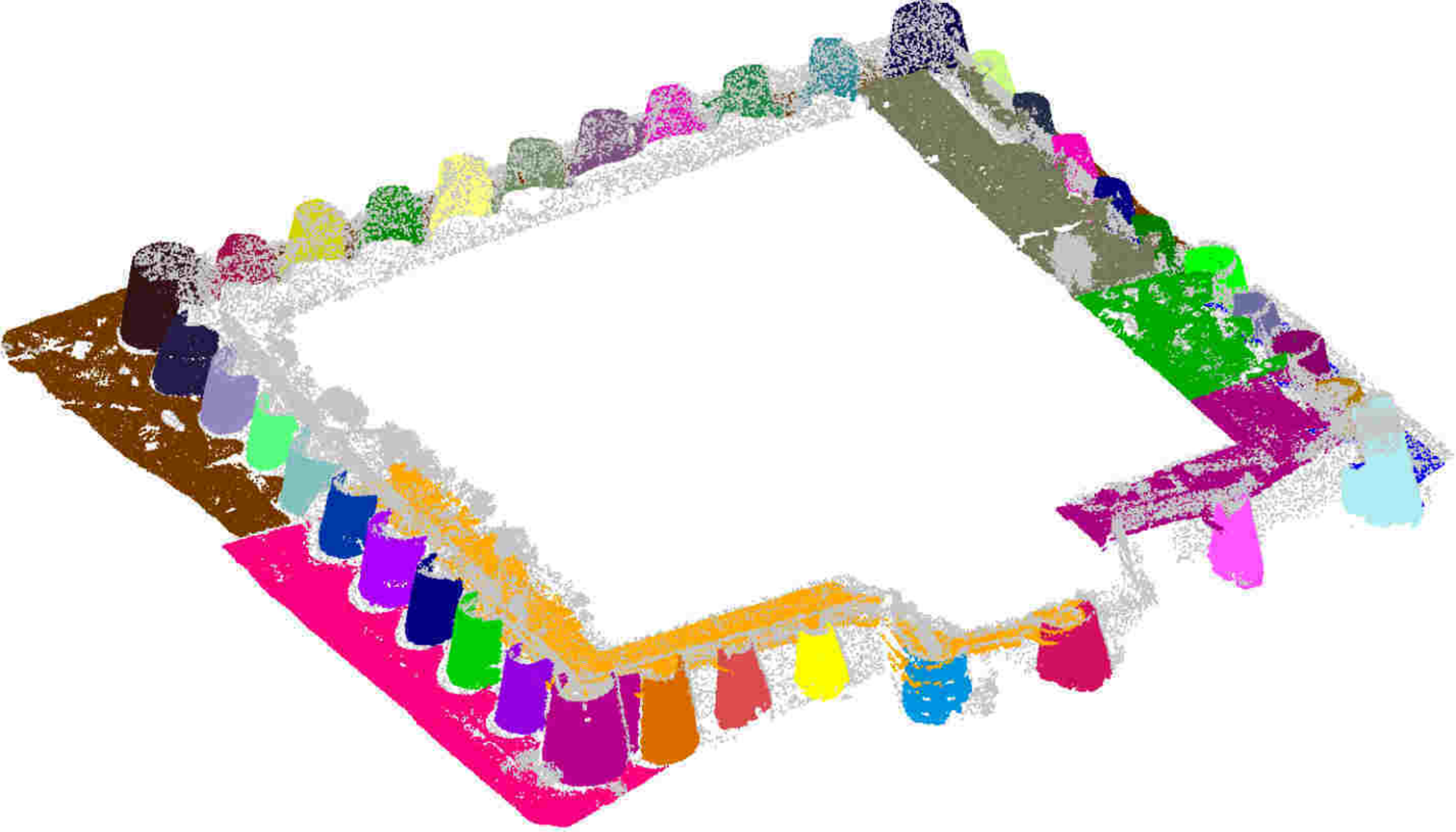}\\
            (a) & (b) & (c) & (d) \\
		\end{tabular}
	} \caption{\label{fig:realresults}
	Sample results showing detailed segmentation. (Row 1) Synthetic Roman building. (Row 2) Masjid Wazir Khan. (Row 3) Derawar fort. (a) Point cloud. (b) Coarse segments obtained using RANSAC~\cite{schnabel-2007-efficient}. (c) DBScan clustering. (d) Our hierarchical segmentation.}
    \end{center}
\end{figure*}

In our experiments, we have successfully segmented two entire buildings which were taken from 3D Warehouse. The temple like building shown in Figure \ref{fig:temple_detail} contained $0.5$M points which were segmented and then our hierarchical tree further divided the building into $10$ segments out of which only $8$ can be categorized as useful. The building shown in Figure \ref{fig:realresults} row 1, is a Roman building. It contained $2.5$M points and $79$ primitives, which were first used as an input to Schnabel's algorithm which segmented the entire building into $24$ planes and $22$ cylinders as shown in Figure \ref{fig:realresults} row 1c. Useful segments, after passing through the DBScan algorithm, as shown in Figure \ref{fig:realresults} row 1c, were given as input to our segmentation algorithm, which produced all $77$ primitives successfully as shown in Figure \ref{fig:realresults} row 1d.

\subsection{Experiment: Derawar Fort}
The real point cloud of Derawar fort, shown in Figure \ref{fig:realresults} row 3a, contained $0.43$B points and $38$ minarets. Initially a relatively low resolution point cloud is fed into  Schnabel's algorithm which segments raw point cloud data containing $38$ primitive shapes into $45$ cylindrical, $29$ planar and $49$ unassigned segments. Figure \ref{fig:realresults} row 3b shows the output of Schnabel's algorithm. Due to over-segmentation this output is  clustered using the DBScan algorithm, as shown in Figure \ref{fig:realresults} row 3c, and sent as an input to our segmentation algorithm. Our algorithm successfully segments all $38$ minarets.

\subsection{Experiment: Masjid Wazir Khan}
The real point cloud of Masjid Wazir Khan, containing $0.3$B points and $12$ domes, $4$ balconies, $6$ minarets, $14$ main arches and $32$ small arches located on the minarets has first been pre-processed to obtain a relatively low resolution point cloud. Initially, Schnabel's RANSAC based algorithm segments the raw point cloud data containing $68$ primitive shapes and returns $12$ planar, $7$ cylindrical and $4$ spherical segments. Schnabel's output is clustered using the DBScan algorithm. These clusters are then interactively selected as input to our segmentation algorithm. Our algorithm successfully recognized and classified $12$ domes, $6$ balconies, $4$ minars and $13$ arches as individual components. Figure \ref{fig:realresults} row $2$ shows the output of our pipeline. 


\subsection{Experiment: Masjid Khudabad }
Masjid Khudabad's real point cloud data contains $0.54$B points out of which we only used 0.64M points. This particular point cloud is mainly defined by $21$ domes and $12$ arches as the primitives. Schnabel's algorithm generates $15$ planar, $4$ cylindrical and $4$ spherical segments as the coarse primitives and their clustered segments are then segmented by our algorithm. Our segmentation algorithm has correctly identified and classified $19$ domes and $7$ out of the $12$ arches that were given to it as the coarse segments.



\begin{figure*}[tbh!]
	\begin{center}\leavevmode \centerline{
		\begin{tabular}{@{}c@{}c@{}c@{}c@{}c@{}}    
				\includegraphics[width=.22\columnwidth]{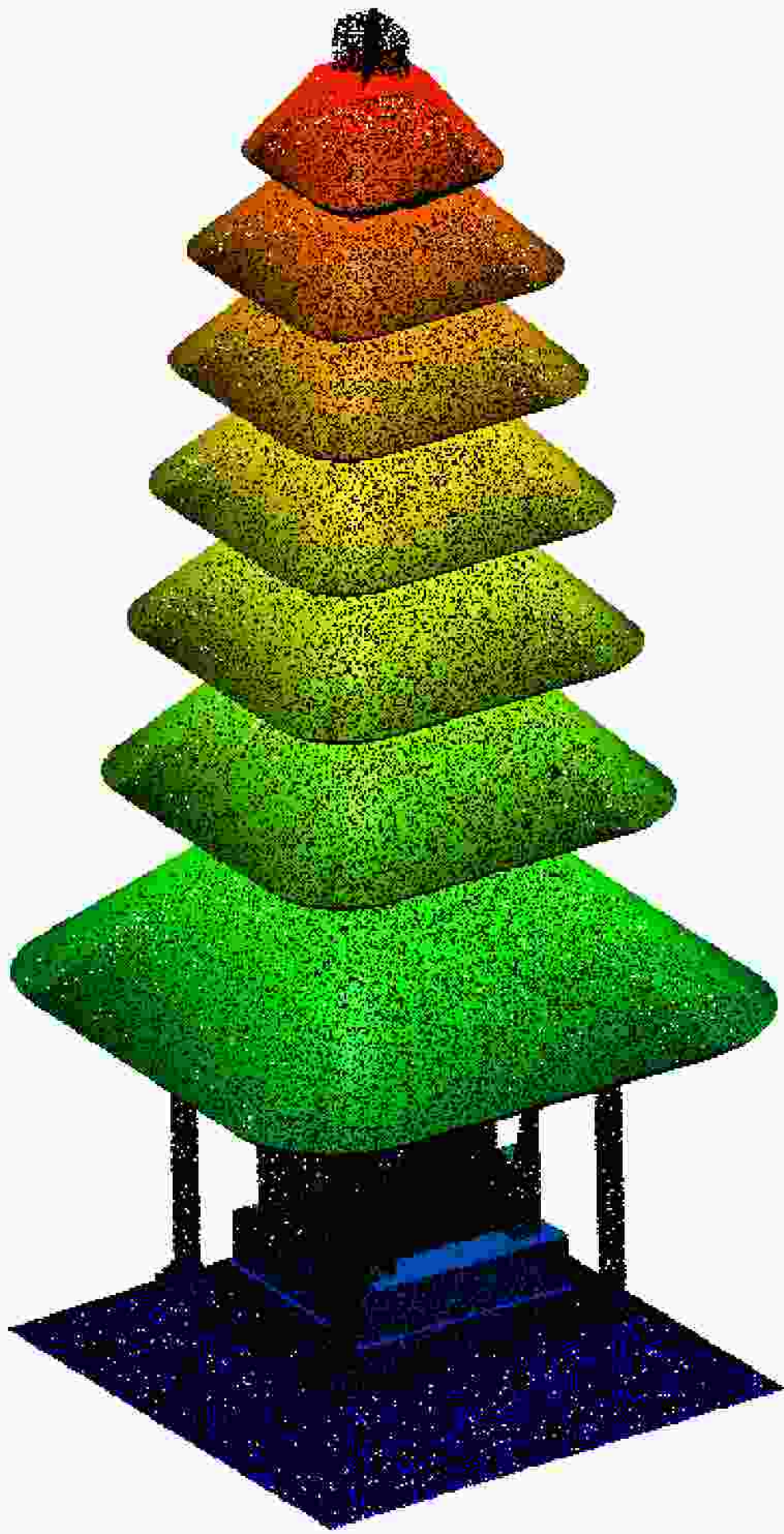} &					
        		\includegraphics[width=.2\columnwidth]{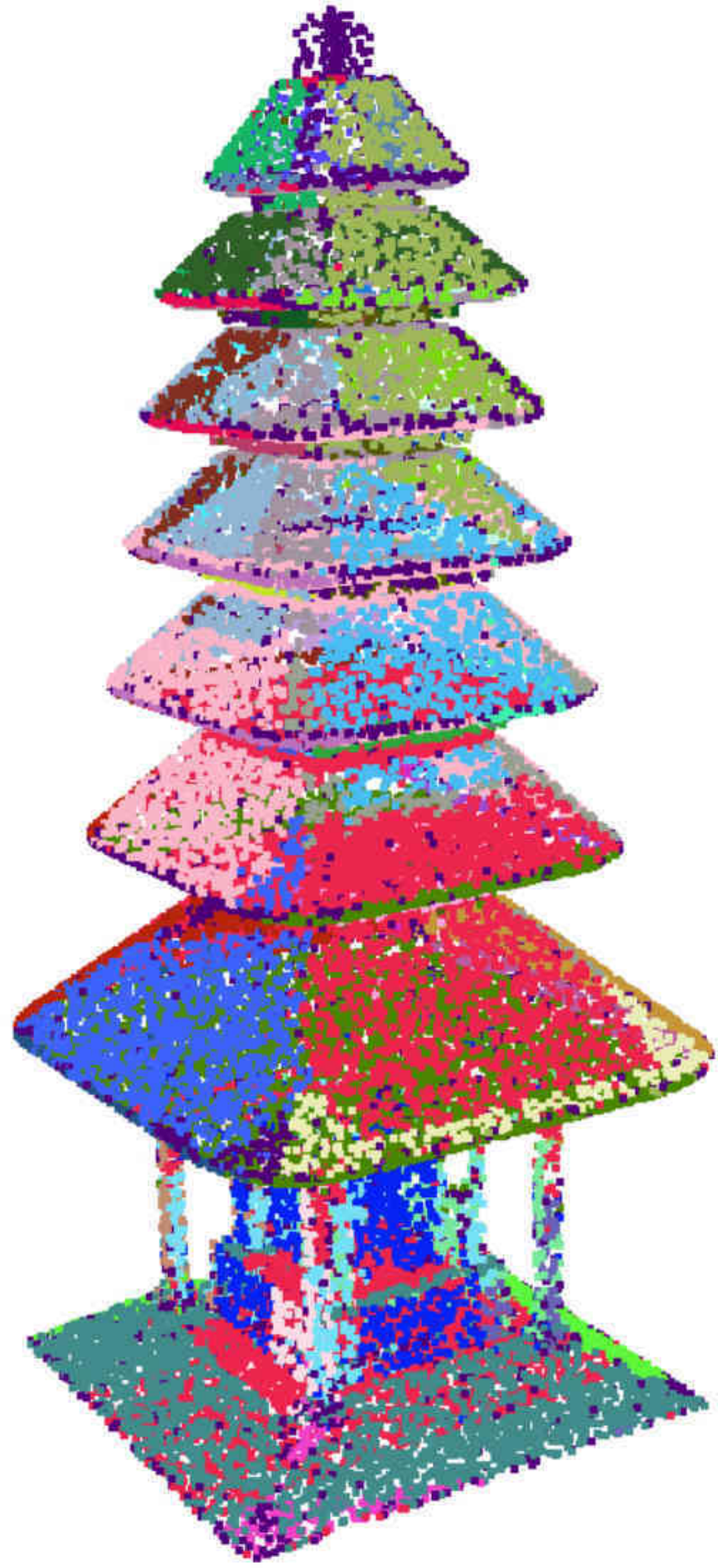} &
        		\includegraphics[width=.22\columnwidth]{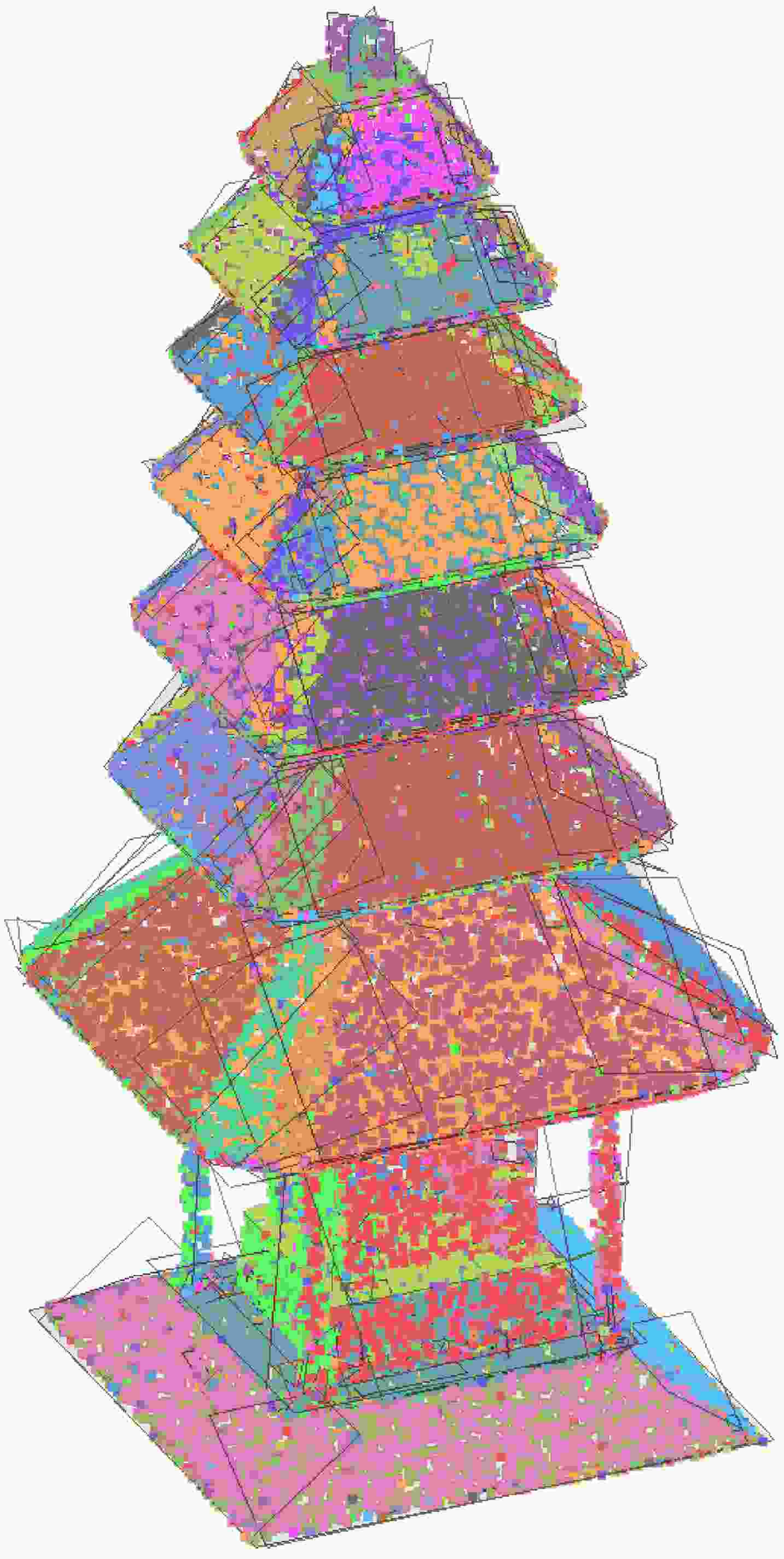} &
				\includegraphics[width=.22\columnwidth]{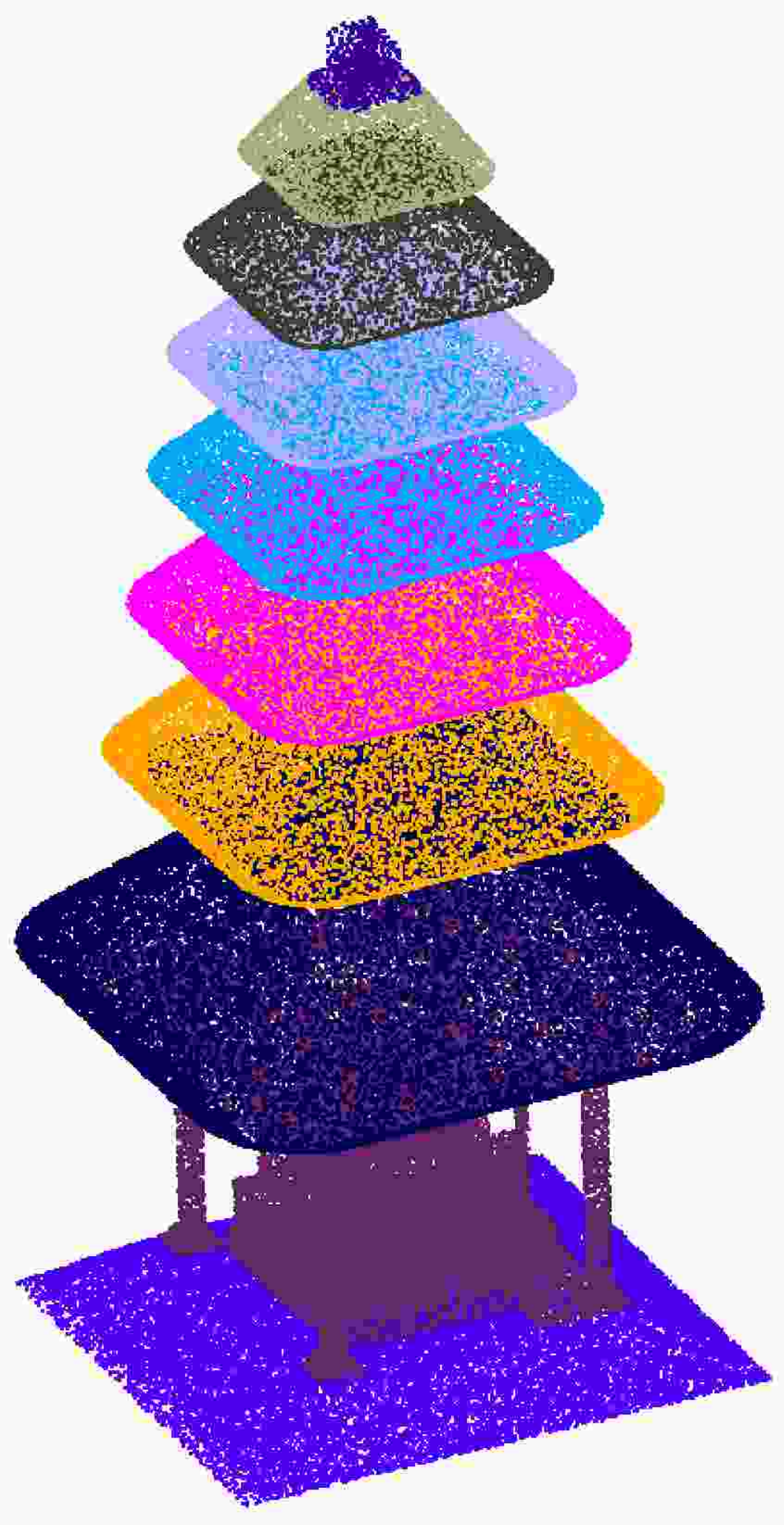} &
				\includegraphics[width=.2\columnwidth]{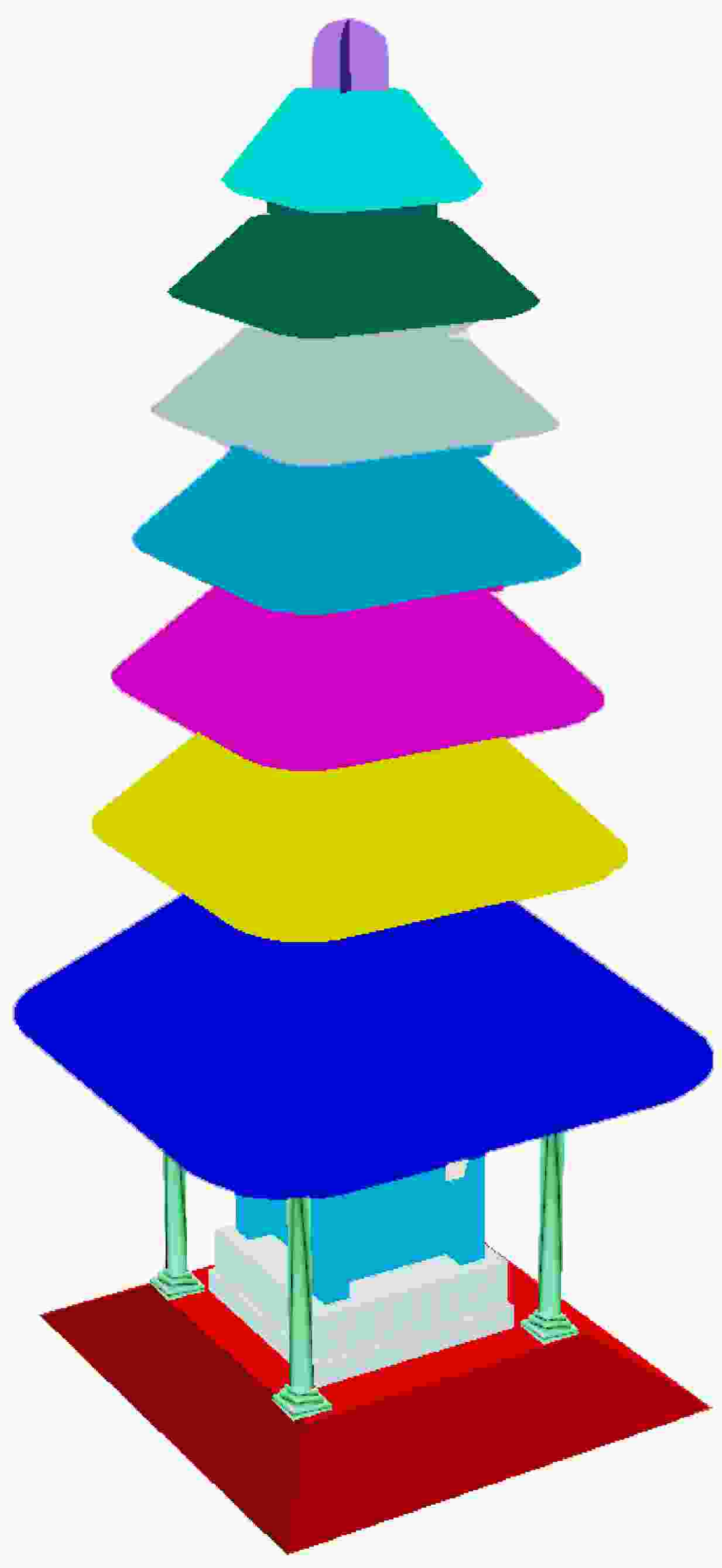}\\				
				\includegraphics[width=.4\columnwidth]{images/coll_original} &	
				\includegraphics[width=.4\columnwidth]{images/colloseum_schnabel3} &
        		\includegraphics[width=.5\columnwidth]{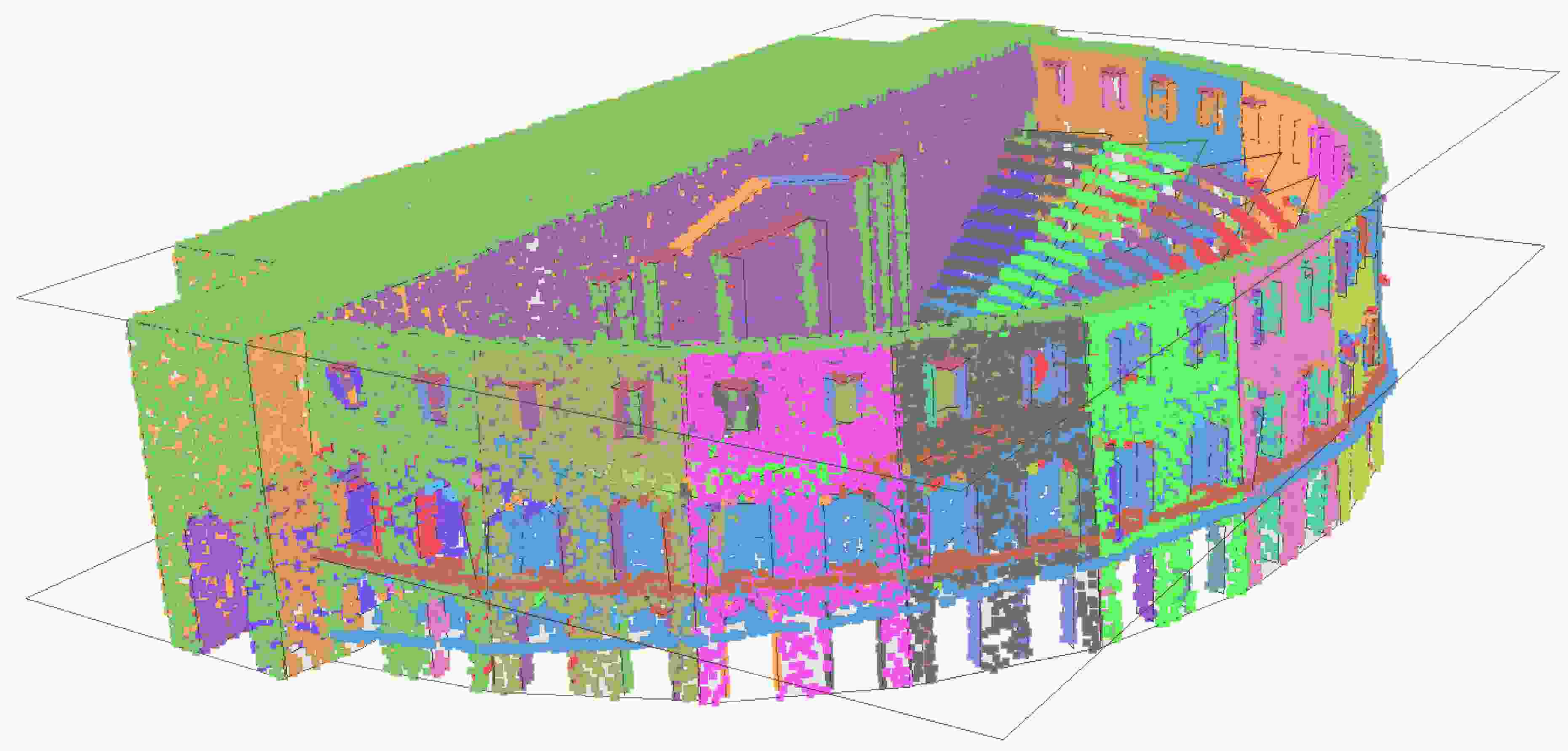} &
				\includegraphics[width=.4\columnwidth]{images/colloseum_output3} &
				\includegraphics[width=.4\columnwidth]{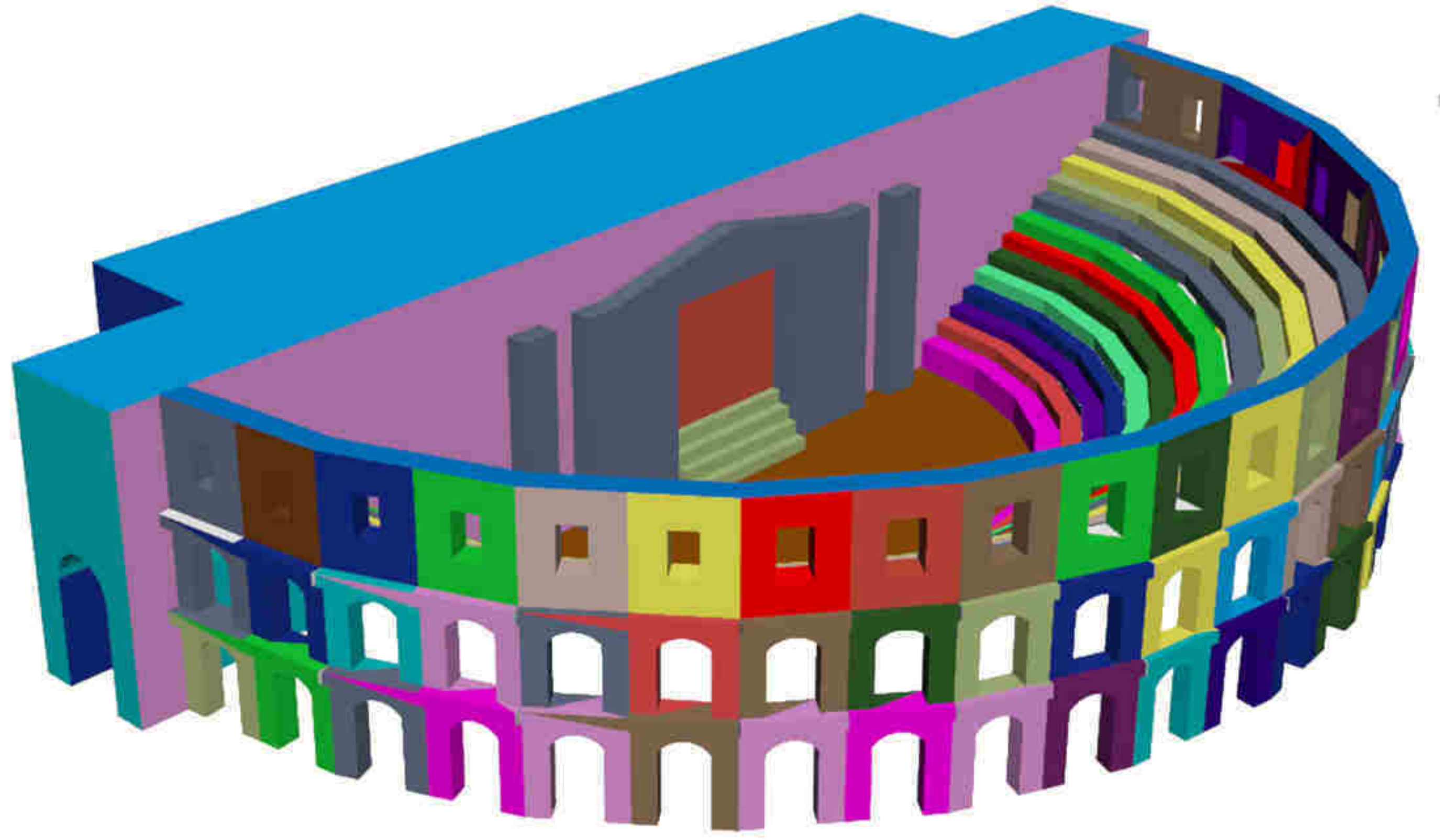}\\										
				(a) & (b) & (c) & (d) & (e)\\
		\end{tabular}
	} \caption{\label{fig:Comparisons} Results from two point cloud datasets showing comparision between proposed hierarchical clustering with regular arrangement of planes (RAPter)~\cite{MonszpartEtAl:RAPter:2015}. (Row 1) Temple. (Row 2) Roman building (a) Point cloud, (b) RANSAC (c) RAPter, (d) ours and (e) ground truth.}
    \end{center}
\end{figure*}

\subsection{Comparisons}
We have compared our hierarchical segmentation algorithm with both plane fitting as well as primitive fitting technique. For plane fitting we have compared with the recently published approach using regular arrangement of planes (RAPter)~\cite{MonszpartEtAl:RAPter:2015}. For primtive fitting technique we compared with the seminal  RANSAC based approach Schnabel\cite{schnabel-2007-efficient}. We have used available implementations of these algorithms and tested them on our data set. Figure \ref{fig:Comparisons}(b) shows the result obtained after applying Schnabel's algorithm on the synthetic point cloud, while Figure \ref{fig:Comparisons}(c) presents the result gathered by using RAPter. Although RAPter clearly dominates in this sceneario, it still fails to segment each individual arch of the curved surface. Our algorithm (~\ref{fig:Comparisons}(d)), in comparison, surpasses both of the aforementioned methods and yields superior segmentation of the curved surface. 

RAPter achieves high accuracy while reconstructing scenes as regular arrangement of planes, therefore, we performed further experiments on planar structures. One such experiment is shown in Figure~ \ref{fig:Comparisons} row 1b. Here, RAPter has successfully fitted planes inside each of the segment thereby reconstructing a segment as a composition of multiple planar surfaces and failing to represent a single segment as one complete entity.  On the other hand, our algorithm successfully separates out each segment as a whole. Figure~\ref{fig:Comparisons} row 1c shows the result of our proposed method. 


\subsection{Timing Performance}
Our input point cloud is first segmented using the model fitting algorithm of Schnabel\cite{schnabel-2007-efficient}. A typical point cloud consists of $800$K to $1200$K points and Schnabel's algorithm produces results in approximately $15-25$ seconds. This segmented output is then clustered using DB-scan algorithm \cite{ester1996density} which takes $0.2$ seconds on each segment and  produces multiple segments based on their spatial location. These clustered segments are then classified by our hierarchical tree algorithm where each projection sequence takes approximately $15$ seconds to perform low level segmentation. It should be noted that these timing numbers are based on unoptimized MATLAB implementation running on core i7 with $32$GB of RAM. 3D convolution network is implemented using an optimized Python code, which first converts a point cloud segment to its Voxel representation using a C++ code in $131$ milliseconds and then takes $50$ milliseconds to classify the 3D Voxel. In contrast to this, the Region growing segmentation algorithm ~\cite{rabbani2006segmentation} typically utilizes $20$ seconds to run on a single segment as shown in figure this.

\section{Conclusions and Future work}
In this work, we have devised a solution for the problem of outdoor 3D scene segmentation and reconstruction. The solution proposed in the form of the hierarchical tree approach is simple but has proved to be effective for the reconstruction of planar and non-planar outdoor scenes. We have successfully used the energy function to explore the 3D data in more detail, while ensuring that we have control over the semi-automatic selection of correct segments. On the other hand, hollow voxels are more informative for 3D data representation and we intend to explore them further in future by increasing the spatial resolution. We also aim to include more semantic information by introducing deep belief network and conditional probabilities in the tree and a richer prior which better encapsulates the semantic information.

This work is applicable to geometry that exhibits structural regularity. Seminal work on structural regularity by Pauly 2008 ~\cite{pauly2008discovering} quantifies regularity in 7 different categories and our projection sequences are applicable to 4 of them: Rot, Trans, RotxTrans and TransxTrans. If coarse segmentation successfully detects diagonal structures/beams, proposed projection sequences can be used after performing axis alignment of the structure. Otherwise, such structures are not segmented further. We take it as a pointer for our future work.



\bibliographystyle{eg-alpha-doi}
\scriptsize

\bibliography{egbibsample}

\end{document}